\documentclass[sigconf]{acmart}

\usepackage{booktabs}
\usepackage{graphicx}
\usepackage{balance}
\usepackage{color,xcolor}
\usepackage{array}
\usepackage{float}
\usepackage{eso-pic}
\usepackage{amsmath}
\usepackage{multirow}
\usepackage{siunitx}
\usepackage{subfigure}
\usepackage[ruled,vlined,linesnumbered]{algorithm2e}

\SetKwComment{Comment}{$\triangleright$\ }{} 
\SetKwInOut{Input}{Require}      
\SetKwInOut{Output}{Ensure}      

\definecolor{FlickrPink}{rgb}{1.00,0.00,0.52}

\newcommand{\NN}{\mathbb{N}}
\newcommand{\RR}{\mathbb{R}}
\newcommand{\itemset}{\Omega}
\newcommand{\singleitem}{\omega}
\newcommand{\recvalue}{\widehat{y}}
\newcommand{\recmod}{\boldsymbol{\recvalue}}
\newcommand{\eval}{\mathcal{R}}
\newcommand{\history}{H}
\newcommand{\symalgo}{\mathcal{A}}

\DeclareMathOperator{\PRR}{PRR}

\DeclareMathOperator*{\argmin}{arg\,min}

\newenvironment{compactitem}
 {\begin{itemize}
    \setlength{\itemsep}{0pt}
    \setlength{\parskip}{0pt}
    \setlength{\parsep}{0pt}
    \setlength{\leftmargin}{1.5em}
 }
 {\end{itemize}}

\AtBeginDocument{%
  }

\setcopyright{none}
\renewcommand{\shortauthors}{}
\acmConference[]{}{}{} 
\acmBooktitle{}        
\acmDOI{}              
\acmISBN{}             
\acmYear{}             
\copyrightyear{}       

\makeatletter
\renewcommand{\footnotetextcopyrightpermission}[1]{}
\makeatother

\begin{document}

\title[What Data is Really Necessary?]{What Data is Really Necessary? A Feasibility Study of Inference Data Minimization for Recommender Systems}

\author{Jens Leysen}
\email{jens.leysen@uantwerpen.be}
\orcid{0000-0003-2561-0932}
\affiliation{%
  \institution{University of Antwerp}
  \city{Antwerp}
  \country{Belgium}
}

\author{Marco Favier}
\email{marco.favier@uantwerpen.be}
\orcid{0000-0003-4676-5896}
\affiliation{%
  \institution{University of Antwerp}
  \city{Antwerp}
  \country{Belgium}
}

\author{Bart Goethals}
\email{bart.goethals@uantwerpen.be}
\orcid{0000-0001-9327-9554}
\affiliation{%
  \institution{University of Antwerp}
  \city{Antwerp}
  \country{Belgium}
}
\affiliation{%
  \institution{Monash University}
  \city{Melbourne}
  \country{Australia}
}
\orcid{0000-0001-9327-9554}

\renewcommand{\shortauthors}{} 

\begin{abstract}
Data minimization is a legal principle requiring personal data processing to be limited to what is necessary for a specified purpose. Operationalizing this principle for recommender systems, which rely on extensive personal data, remains a significant challenge. This paper conducts a feasibility study on minimizing implicit feedback inference data for such systems. We propose a novel problem formulation, analyze various minimization techniques, and investigate key factors influencing their effectiveness. We demonstrate that substantial inference data reduction is technically feasible without significant performance loss. However, its practicality is critically determined by two factors: the technical setting (e.g., performance targets, choice of model) and user characteristics (e.g., history size, preference complexity). Thus, while we establish its technical feasibility, we conclude that data minimization remains practically challenging and its dependence on the technical and user context makes a universal standard for data `necessity' difficult to implement.
\end{abstract}



\maketitle
\pagestyle{plain}

\AddToShipoutPictureBG*{
  \AtPageLowerLeft{
    \put(52.5pt, 40pt){
      \parbox[b]{\columnwidth}{\footnotesize
        \textcopyright~2025 Copyright held by the owner/author(s).
        This is the author's version of the work. It is posted here for your personal use.
        Not for redistribution. The definitive Version of Record was published in
        \textit{Proceedings of the 34th ACM International Conference on Information and
        Knowledge Management (CIKM '25), November 10--14, 2025, Seoul, Republic of Korea},
        \url{https://doi.org/10.1145/3746252.3761058}.
      }%
    }%
  }%
}

\section{Introduction}\label{sec:introduction}
The increasing reliance on data-driven systems has prompted international data protection frameworks, notably the European Union's GDPR~\cite[Art.~5]{GDPR2016}, California's CPRA~\cite{CPRA2020}, Brazil's LGPD~\cite{LGPD2018}, China's PIPL~\cite[Art.~47]{PIPL_Stanford_2021}, and many others~\cite{UKGDPR18, PIPEDA00, SwissFADP20, PIPA11, POPIA13}, to mandate the principle of \textit{data minimization}. 
GDPR Article 5(1)(c), for instance, requires personal data processing\footnote{Under the GDPR, `processing' means `any operation or set of operations which is performed on personal data or on sets of personal data, whether or not by automated means'. 
Examples include collection, storage, and use~\cite[Art.~4(2)]{GDPR2016}.} 
to be `limited to what is necessary' for the specified purpose~\cite[Art.~5]{GDPR2016}. 
Fundamentally, by limiting the volume of personal data that is collected, used, shared, and stored by online services, data minimization aims to reduce the potential for any misuse or privacy breaches, aligning with `privacy by design' principles~\cite{danezis14, cavoukian10}. 
It serves as a proactive strategy, attempting to limit the impact of privacy and security issues \textit{before} they happen.

This legal principle presents significant challenges for modern recommender systems, which typically leverage extensive personal data—primarily behavioral (e.g., clicks, views, likes, purchases) as well as contextual information (e.g., time and location)—to generate personalized recommendations. Operationalizing this legal mandate for recommender systems remains a significant, largely unsolved problem~\cite{finck21}. Prior work suggests that further research is required to~\cite{biega20,finck21}: (1) investigate how data `necessity' can be operationalized, (2) develop technical minimization strategies, and (3) systematically assess both feasibility and practicality.

Regarding this crucial first step of operationalizing `necessity', related work~\cite{rendle23,biega20} has employed formulations for data minimization based on selecting subsets of a uniform size $k$ for all users. We term these approaches \textit{size-constrained minimization}.
We argue, however, that such size constraints may not optimally fulfill the legal principle, primarily because we posit that data \textit{necessity} is \textit{user-dependent}. This dependency arises from factors like individual preference complexity and varying modeling difficulty for different users. In other words, different users likely require different amounts of data to achieve adequate recommendation quality.
Therefore, we introduce \textit{performance-constrained minimization}. This formulation, instead of fixing data size, focuses on achieving a target performance level: it aims to find the smallest data subset for each individual user that satisfies a specified relative performance requirement.

Regarding the aforementioned needs for developing minimization strategies and also assessing feasibility and practicality, this paper conducts a feasibility study specifically for our \textit{performance-constrained} formulation. To scope this study, we focus on the common setting of \textit{implicit feedback data}~\cite{Jannach2018,hu08,joachims07,Rendle09}. We specifically address \textit{inference data minimization}—reducing user data post-training, a focus shared with related work~\cite{biega20, rendle23}. 

This analysis is guided by key questions regarding the performance/size trade-off (RQ1) and the influence of user factors (RQ2):
\begin{enumerate}
\item[\textbf{RQ1}] \textbf{How does recommendation performance relate to data size for various minimization settings?}
\item[\textbf{RQ2}] \textbf{What is the influence of user factors (esp. history size) on data reduction and computational cost?}
\end{enumerate}
The main contributions of this work are:
\begin{compactitem}
    \item A \textbf{novel problem formulation} (Section~\ref{sec:problems}) for dataset minimization that operationalizes 'necessity' by accounting for \textit{per-user performance} requirements, distinguishing our approach from prior formulations (discussed in Section~\ref{sec:background}).
    \item A \textbf{comprehensive evaluation of different minimization algorithms} under our formulation, which includes assessing existing approaches and introducing a \textbf{novel Greedy Removal (GR) variant} (Section~\ref{sec:algorithms}) that achieves superior data reduction under strict performance constraints.
    \item A \textbf{novel evaluation setup} (Section~\ref{sec:experiments}) that, to our knowledge, is the first to employ user stratification for a more nuanced analysis of minimization feasibility.
    \item A \textbf{systematic analysis of key factors} (Section~\ref{sec:discussion}) influencing minimization feasibility, potential, and cost. This analysis reveals that outcomes are critically shaped by technical choices (e.g., recommendation model, performance targets) and user/data characteristics (e.g., users with larger histories often permit greater relative reduction, while others are inherently harder to minimize, likely due to preference complexity).
\end{compactitem}
\section{Background and Related Work}\label{sec:background}
Minimization practices have recently gained attention in the recommender systems research community~\cite{biega20, finck21, shanmugam22, rendle23, niu23, sachdeva22OnSampling, eichinger23, eichinger24}. \citet{finck21} have discussed the difficulties in meaningfully implementing and interpreting legal principles like data minimization and purpose limitation for personalization, profiling, and decision-making systems. Their work, along with other studies~\cite{biega20}, argue for continued research into: (1) operationalizing data `necessity', (2) developing and evaluating technical minimization strategies, and (3) systematically assessing both feasibility and practicality. For further details related to legal interpretations, current obstacles and the far-reaching implications of data minimization for data-driven systems, we refer the interested reader to the techno-legal work by~\citet{finck21}.

Most closely related to our work are the studies by ~\citet{biega20} and ~\citet{rendle23}. Biega et al. provided foundational work by operationalizing the legal principle of data minimization for personalization algorithms~\cite{biega20}. Their key conceptual contribution was proposing \textit{performance-based data minimization} for \textit{inference data}, linking the necessity of data points directly to global model performance. The specific formulations they explored, in the context of \textit{rating data}, notably relied on selecting subsets of a uniform, fixed size $k$ applied across all users while accounting for a shared (global) minimal performance level~\cite{biega20}.
Similarly, the study by Rendle and Zhang~\cite{rendle23} also focuses on reducing \textit{inference data} but for \textit{implicit feedback} recommenders. While not explicitly invoking legal `data minimization' principles, their approach aims to select an interaction subset $I\subseteq H$ of a uniform size $k$ (where $\lvert I\rvert \leq k$) for all users that maximizes expected recommendation performance. They contributed specific greedy (beam) selection algorithms to solve this subset selection task~\cite{rendle23}. 

We categorize both these prior works~\cite{biega20,rendle23} as \textit{size-constrained minimization} formulations. The primary limitation of such uniform size approaches lies in their inability to adapt to varying per-user requirements inherent in the concept of necessity. It neglects that the answer to \textit{"What data is really necessary?"} depends on the characteristics of individual users, such as their preference complexity and the varying difficulty of modeling them accurately.

We briefly discuss more loosely related work. Data minimization has also been explored in more niche settings like decentralized or mobile-to-mobile recommender systems~\cite{eichinger23, eichinger24}. Differing from our focus on already collected data, Shanmugam et al.~\cite{shanmugam22} have concentrated on minimizing data \textit{during collection}. Furthermore, related ideas can be placed under the broader umbrella of minimization strategies for recommender systems~\cite{niu23, larson2017, sachdeva22OnSampling}, including work on concept drift and temporal dynamics~\cite{verachtert2022forgetting, vinagre2012forgetting}, where removing older data can be viewed as a minimization strategy~\cite{verachtert2022forgetting, vinagre2012forgetting}.

Finally, to clearly position our work, we note that this paper focuses on \textit{inference data minimization}—reducing users' histories post-training (akin to~\cite{biega20, rendle23}). It is important to distinguish this from \textit{training data minimization}. Both are vital and complementary research areas for operationalizing the legal principle of data minimization, one focused on minimizing data for generating recommendations (inference) and the other on minimizing data for model building (training). These distinct contexts each merit dedicated discussions and will most likely require different operationalizations and strategies.
\section{Preliminaries}\label{sec:preliminaries}
Let $\Omega=\{\singleitem_1,\dots, \singleitem_n\}$ denote the set of $n$ possible items users can interact with\footnote{The notation used in this paper is primarily based on that of Rendle \& Zhang~\cite{rendle23}.}. We represent the interaction history of a specific user as $H \subseteq \Omega$, which is the set of items the user has previously interacted with (e.g., clicked, purchased, watched). When necessary to distinguish between users, we will denote the history of a specific user $u$ as $H(u)$. In this implicit feedback setting, each interaction is treated as a binary event. With a slight abuse of notation, we will interchangeably consider a set of interactions $H \subseteq \itemset$ and its representation as a binary characteristic vector $H \in \{0,1\}^{n}$.

A \emph{recommendation model}, denoted by $\recmod$, is a function that takes the history of a user $H \subseteq \itemset$ as input and outputs a \emph{ranking}: 
\begin{align*}
    \recmod\colon \{0,1\}^{n}&\to \text{Sym}(n)\\ 
    H&\mapsto 
    \sigma_H
\end{align*}
where $\text{Sym}(n)$ is the set of all possible orders on $n$ elements and $\sigma_H$ is the ranking that given an item $\singleitem_i\in \Omega$ return its position according to the ranking $\sigma_H(\omega_i) = j$. This ranking $\sigma_H$ is used to rank items for actual recommendations, where more relevant items are assumed to have a higher position.

To evaluate the quality of the ranking produced using history $\history$, a \emph{ranking metric} $\eval$ is commonly used. Conceptually, we are interested in the probability $P(\singleitem\mid\history)$ that a user with history $H$ would interact with (e.g. click, buy, watch) a new item $\singleitem$ in the future. Formally, a ranking metric is then defined as:
\begin{equation*}\label{rankingmetric}
\eval_{\history}(\recmod(\history)):=
\sum_{\singleitem\in \itemset\smallsetminus \history}P(\singleitem\mid\history)\varphi\big(\sigma_H(\singleitem)\big)
\end{equation*}
where $\varphi\colon \NN\to\RR$ is a decreasing \emph{reward function}. Hence, a ranking metric reaches its maximum when the items most likely to be interacted with by the user are ranked highest, indicating that the model assigns the highest positions to the most relevant items. Examples are discounted cumulative gain (DCG) or mean reciprocal rank (MRR)\cite{jarvelin02, ricci15, valcarce20}, which differ in their choice of specific \emph{reward function} (also called \emph{discount function}). We use the shorthand notation $\eval_{\history}$ to make explicit that the ranking metric depends on the \emph{ground truth} probabilities $P(\singleitem\mid\history)$.

As mentioned previously, we investigate \textit{inference data minimization}. Formally, we observe that the model need not use the entire history $\history$ as input, but instead may use any subset $I\subseteq \history$.
Given any subset $I\subseteq \history$, we can still assign a ranking to the items in $\itemset$ using the output $\sigma_I = \recmod(I)$. Moreover, the ranking can still be evaluated using the ranking metric $\eval_{\history}$, by calculating
\[
\eval_{\history}(\recmod(I)) = 
\sum_{\singleitem\in \itemset\smallsetminus \history}P(\singleitem\mid \history)\varphi\big(\sigma_I(\singleitem)\big)
\]
Intuitively, removing redundant or non-informative interactions from $\history$ should allow the model $\recmod(I)$ to achieve performance $\eval_\history(\recmod(I))$ close to that of the full history $\eval_\history(\recmod(\history))$. The central challenge, therefore, is to identify the smallest subset $I\subseteq\history$ that maintains a predefined acceptable level of performance relative to $\history$; we explore this formalization in the following section (Section~\ref{sec:problems}).
\section{Problem Formulation}\label{sec:problems}
In contrast to related work~\cite{biega20, rendle23}, which has employed formulations for data minimization based on selecting subsets of a uniform size $k$ for all users (which we term \textit{size-constrained minimization}), we adopt a \textit{performance-constrained minimization} approach. We define the formulation as finding the \emph{smallest} subset $I \subseteq H$ that maintains a predefined level of performance relative to using the full history $H$. This formulation offers distinct advantages. Firstly, it does not impose a uniform output size $k$ across all users, allowing the minimized history size $|I|$ to vary based on individual user characteristics and minimization difficulty. Secondly, the performance constraint is defined \textit{relatively}, ensuring that the quality achieved with the minimized subset $I$ is compared against the quality achieved with the original full history $H$ for that specific user, tackling the challenge of maintaining a consistent level of recommendation performance (relative to the user's own baseline).

Formally, we define this performance-constrained minimization problem for each user as:
\begin{equation}\label{eq:problem_formulation} 
    \begin{aligned}
        \argmin_{ I \subseteq  H} \ \lvert I \rvert
        \quad
        \text{s.t.} \quad \frac{\eval_H\big(\recmod(I)\big)}{\eval_H\big(\recmod(H)\big)} \geq \eta
    \end{aligned}
\end{equation}
The constraint in Equation~\ref{eq:problem_formulation} relies on the \textbf{Performance Retention Ratio (PRR)}, which measures the performance impact of minimization for an individual user. Formally, for a minimization algorithm $\symalgo$ that produces subset $I=\symalgo(H)$:
\begin{equation}\label{eq:prr_definition} 
    \text{PRR}(\symalgo, \mathcal{R}, \recmod, H) := \frac{\mathcal{R}_H(\recmod(I))}{\mathcal{R}_H(\recmod(H))}
\end{equation}
This ratio compares the performance achieved using the minimized subset $I$ to the performance achieved using the full history $H$, based on the chosen ranking metric $\mathcal{R}$ and recommendation model $\recmod$. A PRR value of 1.0 indicates that minimization has preserved the original recommendation performance for that user.

The parameter $\eta \in [0, 1]$ in Equation~\ref{eq:problem_formulation} sets the minimum acceptable threshold for this PRR. An $\eta=1.0$ requires perfect performance retention (PRR = 1.0), while lower values allow for controlled performance degradation (PRR $\geq \eta$) in exchange for potentially smaller subsets $I$.
\section{Minimization Algorithms}\label{sec:algorithms}
To study the relationship between recommendation performance and achievable dataset size for various minimization settings, we identified and implemented several strategies from the related literature~\cite{biega20, rendle23}. We also propose a novel \textbf{Greedy Removal (GR)} variant. We can broadly categorize these algorithms into heuristic and greedy (optimization-based) approaches. 

\subsection{Heuristics}
The heuristic approaches employ simple criteria to select interactions until the performance constraint is satisfied:
\begin{compactitem}
    \item \textbf{Random Selection (RS)}~\cite{biega20, rendle23}: Starting with an empty set, this baseline randomly selects interactions from a user's history.
    \item \textbf{Most Popular (MP) and Least Popular (LP) Selection}~\cite{rendle23}: These baselines select interactions based on the overall popularity of the involved items. The intuition is that more popular items might be more influential, or conversely, that less popular items might represent niche, long-tail user preferences.
    \item \textbf{Embedding Similarity (EmbSim)}~\cite{rendle23}: This heuristic is specifically designed for embedding-based recommendation algorithms. It selects interactions based on the similarity between user and item embeddings.
\end{compactitem}
 
\subsection{Greedy Selection}
The greedy selection algorithms iteratively build a subset of interactions by greedily selecting the interactions that provide the most significant improvement in recommendation performance at each step:
\begin{compactitem}
    \item \textbf{Greedy Forward Selection (GFS)}~\cite{rendle23}: Starting with an empty set, GFS iteratively adds the single interaction from the user's history that results in the largest increase in the chosen ranking metric. 
    \item \textbf{Greedy Beam Forward Selection (GBFS)}~\cite{rendle23}:  GBFS is an extension of GFS that maintains a beam of the top L candidate sets at each iteration. In each step, it evaluates all possible extensions of these candidate sets by adding one more interaction. The top L resulting sets with the best performance become the beam for the next iteration. GBFS explores a wider range of potential subsets compared to GFS.
\end{compactitem}
While forward selection (GFS, GBFS) is intuitive for \textit{size-constrained} problems aimed at maximizing performance for a fixed subset size~\cite{rendle23}, it presents a potential mismatch for our \textit{performance-constrained} objective (Equation~\ref{eq:problem_formulation}). Our goal is to find the \emph{smallest} set $I$ satisfying the performance constraint. GFS, however, is designed to maximize performance gain at each step. When adapted to our formulation (e.g., by adding items until PRR $\ge \eta$), its focus on maximizing immediate gain might lead it to include items that significantly boost performance but are not strictly necessary to meet the threshold $\eta$. This could result in a larger subset $|I|$ than the true minimum required. The observations on the Greedy Forward Selection strategies lead us to define a related approach that is potentially better suited to our performance-constrained objective: Greedy Removal.

\subsection{Greedy Removal}\label{sec:gr}
Given our performance-constrained objective (Eq. 1) and the potential limitations of forward selection methods in finding the \emph{smallest} feasible set, we propose a backward approach: \textbf{Greedy Removal (GR)}. The GR algorithm starts with the full user history $\history$ and iteratively removes interactions. In each step, it identifies the interaction $\singleitem \in I$ considered `least informative'. An interaction is considered least informative if its removal results in the highest possible performance $\eval(\recmod(I \setminus \{\singleitem\}), \history)$, provided that the performance constraint $\PRR \ge \eta$ remains satisfied after its removal. This process continues until no further interactions can be removed without violating the constraint. The detailed procedure is outlined in Algorithm~\ref{alg:gr_compact}.
\begin{algorithm}[tbh]
\caption{Greedy Removal (GR) Algorithm}
\label{alg:gr_compact}
\Input{History $\history \subseteq \itemset$, Model $\recmod$, Metric $\eval$, Threshold $\eta \in [0, 1]$}
\Output{Minimized subset $I \subseteq \history$}

$I \leftarrow \history$, $R_{\history} \leftarrow \eval(\recmod(\history), \history)$

\While{true}{
    $\text{best\_item} \leftarrow \text{null}$ \\
    $\text{max\_perf} \leftarrow -\infty$

    \ForEach{$\singleitem \in I$}{
        $R_{\text{temp}} \leftarrow \eval(\recmod(I \setminus \{\singleitem\}), \history)$ \\
        \If{($R_{\text{temp}} / R_{\history} \ge \eta$) \textbf{and} ($R_{\text{temp}} > \text{max\_perf}$)}{
             $\text{max\_perf} \leftarrow R_{\text{temp}}$ \\
             $\text{best\_item} \leftarrow \singleitem$
        }
    }

    \uIf{$\text{best\_item} \neq \text{null}$}{ 
        $I \leftarrow I \setminus \{\text{best\_item}\}$
    }
    \Else{
        \textbf{break} 
    }
}
\KwRet{$I$}
\end{algorithm}

This backward approach, starting from the full interaction set, offers potential advantages for our performance-constrained formulation. By considering the complete interaction context from the outset, GR evaluates removals within the full context, which might better preserve complex dependencies between interactions compared to forward methods that build the history structure incrementally. This could potentially avoid suboptimal paths arising from early greedy choices in forward selection, possibly leading to smaller final subsets $I$. Furthermore, GR maintains strict feasibility during its search. Since it begins with a valid solution (the full history $H$) and ensures PRR $\ge \eta$ after every potential removal, all intermediate subsets generated during the process are feasible solutions. This ensures the algorithm provides a valid subset meeting the performance requirement even when terminated prematurely, unlike forward selection methods which yield a valid solution only upon reaching the performance threshold.

The suitability of forward versus backward approaches also depends on the target performance level $\eta$. If $\eta$ is low (requiring only minimal performance retention), a forward approach like GFS might be more efficient. GFS explores solutions by iteratively adding items, conceptually searching from an empty set towards the full history (search direction: $\emptyset \rightarrow \history$). It might need only a few high-impact selections to reach a low performance threshold quickly. Conversely, if $\eta$ is high (requiring near-perfect performance retention), a backward approach like GR may be preferable. GR starts with the full, feasible history H and explores solutions by removing items, searching from the full set towards smaller subsets (search direction: $\history \rightarrow \emptyset$). This process of making only the few necessary removals might be computationally cheaper than GFS needing to add nearly all interactions to reach a high threshold.
\section{Experimental Setup}\label{sec:experiments}
To ensure the reproducibility of our findings, we have made our source code, experimental setup, and results openly available in a Github repository.\footnote{\label{fn:repo}\url{https://github.com/Djensonsan/cikm2025-data-minimization}} The repository includes configuration files detailing our data preprocessing, dataset splitting procedures, and hyperparameter tuning. It also contains more comprehensive results than the findings presented in this paper.

\subsection{Datasets}
We experiment with three widely-used, open-source recommendation datasets: \href{https://www.kaggle.com/datasets/netflix-inc/netflix-prize-data}{\textit{Netflix}}, \href{https://grouplens.org/datasets/movielens/}{\textit{MovieLens}}~\cite{harper16}, and \href{http://millionsongdataset.com/}{\textit{Million Song Dataset}} (MSD)~\cite{bertin11}. These datasets are often converted into implicit feedback datasets by binarizing the ratings and play counts~\cite{liang18, steck19_a, steck19_b, shenbin20, rendle22, rendle23}.
We use the default preprocessing steps outlined by Liang et al.~\cite{liang18}. In short, the datasets were filtered for both items and users with minimum activity levels. For rating data, we binarized the values by treating ratings above four as positive interactions, given that the ratings are on a five-point scale. 

\subsection{Recommendation Models}
As previously mentioned, this paper focuses on the common setting of \textit{implicit feedback data}. To scope this study, we select two algorithms from the item-item similarity family: \textit{EASE}~\cite{steck19_a, steck19_b} and \textit{ItemKNN}~\cite{sarwar01}. We use implementations from RecPack~\cite{michiels22}, a comprehensive open-source framework designed for recommender system experimentation. 

These algorithms perform inferences through a straightforward matrix multiplication of a user's interaction history with an item-item similarity matrix, making them fast and efficient. In terms of performance, \textit{ItemKNN} is considered a strong baseline algorithm~\cite{dacrema21, dacrema19}. \textit{EASE}, however, is state-of-the-art on several datasets (\textit{Netflix}, \textit{MSD}, \textit{MovieLens}, \textit{Amazon Digital Music}, and \textit{Epinions}~\cite{steck19_a, steck19_b, shehzad23, dacrema21}), consistently outperforming more complex models, including VAE and DAE architectures.

\subsection{Ground Truth Estimation}
Our problem formulation (see Section \ref{sec:problems}) requires the calculation of a ranking metric $\eval$. In practice, the ground truth probabilities $P(\singleitem\mid\history)$ needed to compute the metric $\eval$ are unknown.   

A common method to address this is \textit{hold-out estimation}, where we partition a user's history $H$ into two disjoint subsets: $H_{out}=H \setminus H_{in}$. This approach approximates $P(\singleitem\mid H_{in})$ using an indicator function based on the held-out set of interactions: $P(\singleitem\mid H_{in}) \approx \delta(\singleitem \in H_{out})$. For individual users, $H_{out}$ can provide a very sparse (and also binary) relevance signal (e.g., only a few items are considered relevant with $P=1$, and all unobserved items as irrelevant with $P=0$). Severe sparsity makes hold-out estimation ill-suited for effectively evaluating the performance for an individual user\footnote{Please note, we are not stating \textit{hold-out estimation} is an invalid approach for calculating aggregate metrics. When averaging over a large number of users their hold-out sets provide a good estimate for the aggregate performance. The issue is that there is no such guarantee for a single hold-out to be a good estimate for an individual user.}.

To address these limitations and obtain a denser, graded estimate of $P(\singleitem\mid \history)$, we adopt \textit{output estimation}, a method that estimates $P(\singleitem\mid\history)$ using the output of the recommendation model $\recmod$ itself, as proposed by~\citet{rendle23}. This method assumes that $\recmod$, trained to predict user interactions, already captures the notion of interaction probability through its output scores or ranks. Therefore, the rank $i$ assigned by $\recmod$ (given an input history H) to items $\omega \notin \history$ can be transformed into an estimate of $P(\singleitem\mid H)$ using a parametric function, such as $f_{K, \gamma}(i)$~\cite{rendle23}:
\begin{equation} \label{eq:f_gamma_k} 
  f_{K, \gamma}(i):=\delta(i \leq K)\frac{(i+1)^{\gamma} - (K+1)^{\gamma}}{2^{\gamma} - (K+1)^{\gamma}}
\end{equation}
where $K\in\NN$ is a ranking cut-off parameter and $\gamma\in\RR$ controls the decay rate. The probability $P(\singleitem\mid\history)$ is then approximated by the output of this function applied to the item ranks generated by the model $\recmod(\history)$:
\[
  P(\singleitem\mid\history) \approx f_{K, \gamma}(\sigma_{H}(\singleitem))
\]
While still an approximation, a key advantage of using $f_{K, \gamma}(i)$ is the ability to control the density of the evaluation target, which is not the case for \textit{hold-out estimation}. The ranking cut-off $K$ allows us to define how many top-ranked items receive non-zero probability (relevance) estimates, thus generating a denser and more graded signal compared to the binary indications from a hold-out set. Therefore, for estimating $P(\singleitem\mid H)$ in our evaluation of $\eval$, we adopt this parametric approach via $f_{K,\gamma}$. For a comprehensive analysis of the characteristics and benefits of this \textit{output estimation}, we refer the reader to~\cite{rendle23}.

\subsection{Dataset Splitting}\label{splitting}
\begin{table}
    \caption{Statistics of the benchmark datasets after preprocessing. 
    }
    \label{tab:stats}
    \centering
    \begin{tabular}{llll}
        \toprule
        \textbf{Statistic} & \textbf{MovieLens25M} & \textbf{Netflix} & \textbf{Million Song} \\
        \midrule
        \# Users           & 160,770         & 463,435          & 571,355\\
        \# Items           & 19,937          & 17,769          & 41,140\\
        \# Interactions    & 12.4M           & 56.9M           & 33.6M\\
        Inter. Density     & 0.39\%          & 0.69\%          & 0.14\%\\
        \midrule
        \# Val. Users (Rec.)      & 10,000           & 40,000           & 50,000 \\
        \# Val. Users (Est.)      & 1,000           & 1,000           & 1,000 \\
        \# Test Users      & 10,000           & 40,000           & 50,000 \\
        \bottomrule
    \end{tabular}
\end{table}
In our study, we employ the \textit{Strong Generalization} setup for dataset splitting, widely utilized to assess the effectiveness of recommender systems~\cite{liang18, rendle23,steck19_a,shenbin20,rendle22,kim19,lobel20}. This method randomly allocates users to either the training, validation, and test segments, such that there is no overlap in users between segments. Within the validation and test segments, 80\% of a user's interactions form the \textit{fold-in} set, used as input for the recommendation model, while the remaining 20\% constitute the \textit{hold-out} set, serving to evaluate the model's performance through various metrics. Rendle \& Zhang~\cite{rendle23} extended this method by introducing an additional validation split to optimize the ground truth estimation for the dataset minimization procedure. An overview of the dataset statistics and splitting approach is detailed in Table \ref{tab:stats}. In summary, we use the following splits:
\begin{compactitem}
  \item \textbf{Train Split.}
  The train split is used for model fitting.
  \item \textbf{Validation Split - Recommendation.} The validation split is employed to tune the hyperparameters of the recommendation algorithms, such as the L2-norm regularization of the EASE algorithm. NDCG@100 is used as the optimization metric.
  \item \textbf{Validation Split - Ground Truth Estimation.}
  The function $f_{\gamma, K}(t)$, defined as the ground truth estimator in Equation~\ref{eq:f_gamma_k}, depends on two hyperparameters, $\gamma$ and $K$. We optimize these hyperparameters on a separate validation split of 1,000 users, adopting the same grid search ranges for $\gamma$ and $K$ as~\citet{rendle23}.
  \item \textbf{Test Split.}  To evaluate the minimization procedure, we minimize the \textit{fold-in} set for each user based on its estimated ground truth. The \textit{hold-out} set for each user is then used to compare the performance of the minimized dataset against the performance of the full dataset across various metrics (e.g., NDCG, Recall in Table \ref{tab:minimizer_test_eval}).
\end{compactitem}

\subsection{Dataset Stratification}\label{sec:stratification}
Our second research question (\textbf{RQ2}) investigates how user characteristics influence data minimization potential and cost. We hypothesize that user history size, $|H(u)|$, is a primary characteristic impacting both the achievable data reduction and the computational efficiency of minimization algorithms. To specifically analyze the role of this factor, we employ stratification based on user history size. This approach allows us to move beyond global averages, and examine how minimization effectiveness varies across user groups defined by the size of their interaction history.

Our analysis focuses on understanding trends across the typical range of user history sizes. To achieve this and mitigate the influence of extreme outliers (users with exceptionally long histories), we first identify the 95th percentile of history size for each dataset. The analysis and visualization then concentrate on the users within this range (from 0 interactions up to the 95th percentile value). This range is divided into five equal-width bins based on history size. Users with histories longer than the 95th percentile are excluded from this specific stratified analysis to ensure that trends across the main distribution are not distorted by potentially anomalous behavior associated with extreme history sizes.

Consequently, the number of users per bin can vary substantially, reflecting the natural distribution of history sizes within each dataset. Specifically, the Netflix dataset features a higher proportion of users with long histories compared to the Million Song Dataset, which generally has the shortest interaction histories among those studied. For each resulting stratum (bin), we will analyze and report key evaluation metrics separately.

\subsection{Evaluation Metrics}\label{metrics}

Let $\symalgo(H)$ be the minimization algorithm that given history $H$ returns a minimized set $I=\symalgo(H)$. To evaluate the different facets of the data minimization process, capturing both effectiveness (data reduction) and efficiency (computational cost), we consider the following key metrics:

\begin{compactitem}
    \item \textbf{Minimization Ratio (MR)}: For a set of users $U$, MR is defined as:
    \[
        \text{MR}(\symalgo, U) := \frac{\sum_{u \in U} |I(u)|}{\sum_{u \in U} |H(u)|} \quad \text{where } I(u)=\symalgo(H(u)).
    \]
    This metric compares the size of the minimized histories to the size of the original histories (for a single user or set of users). A lower MR indicates greater data reduction. 
    \item \textbf{Sample Efficiency (SE)}: SE is defined as the number of recommendation model ($\recmod$) inferences required by algorithm $\symalgo$ to produce the minimized set $I$ for a given history $H$. Since the evaluated algorithms involve repeated model queries, and model inference is typically the dominant cost, SE provides a measure of the minimization algorithm's computational complexity that is independent of specific hardware, implementation details, or parallelization strategies.
\end{compactitem}

Together, MR and SE evaluate the trade-off between the amount of data reduction achieved and the computational cost incurred by the minimization algorithm itself. As a sidenote, for all experiments in this paper, NDCG@100 is adopted as the ranking metric for calculating and evaluating the performance constraints.
\section{Discussion}\label{sec:discussion}
In this section, we discuss our findings in relation to the research questions posed in the introduction, interpreting the empirical results obtained from our experiments. Given the extensive range of settings explored (including different recommendation models, minimization algorithms, $\eta$ values, and datasets), our discussion will necessarily focus on presenting representative results to highlight key insights. For the full, comprehensive set of results, detailed experimental configurations, and accompanying source code, we refer interested readers to our openly available project repository.

Before we discuss the research questions, we briefly validate the ground truth estimation method from~\citet{rendle23} as a strong estimator. Its effectiveness is confirmed by showing that when user histories are minimized using this estimator against a target performance threshold ($\eta$), their empirical Performance Retention Ratio (PRR) on an external test set closely aligns with the targeted performance. For instance, Table~\ref{tab:minimizer_test_eval} demonstrates for $\eta=0.98$ that the test set PRR indeed aligns well with the 98\% target. This alignment consistently holds for other $\eta$ values (1.00, 0.99) and datasets (MSD, Movielens), with full details available in our results repository.

\begin{table}[htbp]
    \caption{Test set performance comparison for minimized user histories on Netflix (target $\eta$ of $98\%$). Shows absolute metrics for the minimized histories (Min) and Performance Retention Ratio (PRR [\%]) for various metrics and minimizers applied to EASE and ItemKNN models. Full history baseline performance is provided in subheaders.}
    \label{tab:minimizer_test_eval}
    \centering
    \begin{tabular}{@{}l cr cr cr@{}}
        \toprule
        Minimizer & \multicolumn{2}{c}{NDCG@100} & \multicolumn{2}{c}{Recall@20} & \multicolumn{2}{c}{Recall@50} \\
        \cmidrule(lr){2-3} \cmidrule(lr){4-5} \cmidrule(lr){6-7}
                  & Min    & PRR (\%) & Min     & PRR (\%) & Min     & PRR (\%) \\ 
        \midrule
        \multicolumn{7}{@{}l}{\textit{EASE w. Full History: NDCG=0.382, R@20=0.345, R@50=0.432}} \\
        \midrule
        EmbSim & 0.377 & 98.5\% & 0.341 & 99.0\% & 0.426 & 98.7\% \\
        GR     & 0.375 & 98.2\% & 0.335 & 97.2\% & 0.419 & 96.9\% \\
        GBFS   & 0.377 & 98.5\% & 0.336 & 97.6\% & 0.422 & 97.6\% \\
        GFS    & 0.377 & 98.6\% & 0.337 & 97.7\% & 0.423 & 98.0\% \\
        LP     & 0.380 & 99.5\% & 0.342 & 99.3\% & 0.430 & 99.5\% \\
        MP     & 0.374 & 97.9\% & 0.339 & 98.3\% & 0.424 & 98.0\% \\
        RS     & 0.379 & 99.1\% & 0.341 & 98.8\% & 0.428 & 98.9\% \\
        \midrule
        \multicolumn{7}{@{}l}{\textit{ItemKNN w. Full History: NDCG=0.324, R@20=0.282, R@50=0.375}} \\
        \midrule
        EmbSim & 0.320 & 98.9\% & 0.280 & 99.4\% & 0.372 & 99.0\% \\
        GR     & 0.317 & 98.0\% & 0.275 & 97.5\% & 0.368 & 98.0\% \\
        GBFS   & 0.320 & 98.8\% & 0.279 & 98.8\% & 0.370 & 98.5\% \\
        GFS    & 0.320 & 98.9\% & 0.279 & 98.7\% & 0.371 & 98.8\% \\
        LP     & 0.327 & 101.1\% & 0.286 & 101.5\% & 0.380 & 101.1\% \\
        MP     & 0.314 & 97.1\% & 0.275 & 97.6\% & 0.367 & 97.7\% \\
        RS     & 0.321 & 99.2\% & 0.280 & 99.1\% & 0.373 & 99.3\% \\

        \bottomrule
    \end{tabular}
\end{table}

\subsubsection*{RQ1: How does recommendation performance relate to data size for various minimization settings?}

We focus first on allowing a small relative performance decrease ($\eta=0.98$). Our results demonstrate the general feasibility of substantial data reduction when perfect performance retention is not required (see Table~\ref{tab:eta_98_comparison}). For the EASE model, the lowest MR values ranged from 49.3\% (Netflix) to 67.3\% (MSD). While for the ItemKNN model, dataset reduction was even more substantial, with the lowest MR values ranging from 15.5\% (MovieLens) to 49.1\% (MSD).

\begin{table}[h]
\caption{Comparison of MR (\%) and SE (\#) for Greedy Algorithms at $\eta=0.98$ across Models and Datasets.}
\label{tab:eta_98_comparison} 
\centering
\begin{tabular}{@{}l l rr rr rr@{}}
\toprule
& & \multicolumn{2}{c}{\textbf{GFS}} & \multicolumn{2}{c}{\textbf{GBFS ($L=5$)}} & \multicolumn{2}{c}{\textbf{GR}} \\
\cmidrule(lr){3-4} \cmidrule(lr){5-6} \cmidrule(lr){7-8}
\textbf{Model} & \textbf{Dataset} & \textbf{MR} & \textbf{SE} & \textbf{MR} & \textbf{SE} & \textbf{MR} & \textbf{SE} \\ 
\midrule
\multirow{3}{*}{EASE}
 & Netflix & 77.3 & 7,084 & 71.5 & 33,126 & \textbf{49.3} & 6,258 \\
 & ML25M   & 75.0 & 2,403 & 68.4 & 10,930 & \textbf{60.8} & 1,896 \\
 & MSD     & 75.4 & 1,229 & 72.3 & 5,658 & \textbf{67.3} & 834 \\
\midrule 
\multirow{3}{*}{ItemKNN}
 & Netflix & 30.1 & 2,703 & \textbf{27.0} & 11,232 & 30.0 & 7,221 \\
 & ML25M   & 18.3 & 485 & \textbf{15.5} & 1,760 & 22.6 & 2,559 \\
 & MSD     & 58.2 & 959 & 54.5 & 4,265 & \textbf{49.1} & 1,079 \\
\bottomrule
\end{tabular}
\end{table}

Focusing next on the strictest requirement---perfect performance retention ($\eta=1.0$)---our results highlight significant challenges and model dependencies (Table~\ref{tab:eta_1_comparison}). For the EASE model, demanding perfect retention proved largely impractical, permitting very little data removal; MR values in Table~\ref{tab:eta_1_comparison} are consistently high (i.e., 97.8\% - 99.8\%) across datasets. Conversely, for the ItemKNN model, more substantial data reduction was possible, particularly for Netflix (92.5\%) and MSD (88.2\%). 

\begin{table}[h] 
\caption{Comparison of MR (\%) and SE (\#) for Greedy Algorithms at $\eta=1.0$ across Models and Datasets.} 
\label{tab:eta_1_comparison} 
\centering
\begin{tabular}{@{}l l rr rr rr@{}}
\toprule
& & \multicolumn{2}{c}{\textbf{GFS}} & \multicolumn{2}{c}{\textbf{GBFS ($L=5$)}} & \multicolumn{2}{c}{\textbf{GR}} \\
\cmidrule(lr){3-4} \cmidrule(lr){5-6} \cmidrule(lr){7-8}
\textbf{Model} & \textbf{Dataset} & \textbf{MR} & \textbf{SE} & \textbf{MR} & \textbf{SE} & \textbf{MR} & \textbf{SE} \\ 
\midrule
\multirow{3}{*}{EASE} 
 & Netflix & 99.8 & 7,619 & 99.7 & 37,315 & \textbf{98.5} & 308 \\
 & ML25M   & 99.8 & 2,646 & 99.7 & 12,764 & \textbf{98.7} & 135 \\
 & MSD     & 99.2 & 1,333 & 98.8 & 6,252 & \textbf{97.8} & 126 \\
\midrule 
\multirow{3}{*}{ItemKNN} 
 & Netflix & 99.8 & 7,619 & 99.5 & 37,305 & \textbf{92.5} & 1,437 \\
 & ML25M   & $\overline{100}$ & 2,646 & $\overline{100}$ & 12,799 & \textbf{99.9} & 58 \\
 & MSD     & 96.9 & 1,330 & 95.9 & 6,245 & \textbf{88.2} & 393 \\
\bottomrule
\end{tabular}
\end{table}

The previous analysis highlights that the target $\eta$ is only one component determining the effective "difficulty" of the minimization task. The actual potential for data reduction is influenced by the interplay between $\eta$ and other factors, including the recommendation model ($\recmod$), the nature of the ground truth ($P(\omega|H)$ proxy), the chosen evaluation metric ($\mathcal{R}$), and its parameters (e.g., the ranking cut-off $K$ in NDCG@K) as well as the specific dataset. Our findings suggest that scenarios with high effective difficulty (e.g., strict $\eta$, sensitive model) can quickly lead to situations where almost no data can be removed, as seen for EASE at $\eta=1.0$ (Table~\ref{tab:eta_1_comparison}). Conversely, scenarios with low effective difficulty can permit significant data removal.

These findings underscore the complex interplay between technical decisions and data minimization. While certain technical parameters, such as the choice of recommendation model, are likely predetermined prior to the minimization process, it's unclear how other technical choices should be decided. Specifically, such choices include the selection of metrics to define performance constraints (e.g., our choice of NDCG@100) and the setting of performance thresholds (e.g., the $\eta$ value). Because these specific technical choices effectively define what data is deemed `necessary,' it significantly complicates efforts to consistently interpret and enforce data minimization mandates. 

Turning to the comparison between minimization algorithms, this choice also significantly impacts the achievable trade-off between data reduction and computational cost. While heuristic methods (RS, LP, MP, EmbSim) are computationally cheaper (lower SE), they generally achieved less data reduction (higher MR) compared to greedy approaches (GFS, GBFS, GR) (see Tables 5 and 6). Among the heuristics, Most Popular (MP) and Embedding Similarity (EmbSim) tended to perform better in terms of data reduction. 

We now compare the greedy minimization algorithms: Greedy Removal (GR), Greedy Forward Selection (GFS), and Greedy Beam Forward Selection (GBFS), focusing first on their effectiveness in data reduction (Minimization Ratio, MR). For stricter performance requirements (e.g., $\eta=1.0$ and $\eta=0.99$), the proposed Greedy Removal (GR) demonstrated the most data reduction across most settings. Under the perfect retention constraint ($\eta=1.0$), GR achieved a minimization ratio of 92.5\% for ItemKNN on Netflix, while GFS and GBFS had minimization ratios above 99.5\%. Similarly, on MSD, GR achieved a minimization ratio of 88.2\%, while GFS and GBFS had minimization ratios above 95.9\%. As previously discussed, the strict performance requirement turned out to be practically infeasible for the EASE model across datasets, although GR still generally achieved the largest reduction possible. A reason for this is that by considering the complete interaction context from the outset, GR evaluates removals within the full context, which likely better preserves complex dependencies between interactions compared to forward methods that build history structure incrementally. 

However, as the performance threshold is relaxed (e.g., $\eta=0.98$, Table \ref{tab:eta_98_comparison}), the relative minimization ratios shift. For the EASE model, GR still obtains the lowest minimization ratios, with results between 49,3\% and 67.3\% across the different datasets. GBFS and GFS comparatively have minimization ratios between 68.4\% and 77.3\%. For the ItemKNN model, however, relaxing the threshold makes the forward selection methods (GFS, GBFS) notably more competitive against GR in terms of data reduction. In certain scenarios, they even achieve superior minimization. A clear illustration comes from the MovieLens dataset: here, GFS reduced the data to just 18.3\% of its original size, and GBFS performed even better, achieving a 15.5\% minimization ratio. For this same dataset, the backward GR approach resulted in a comparatively larger subset with an MR of 22.6\% (Table 3). This aligns with the intuition from Section~\ref{sec:algorithms} that forward search can be efficient for lower $\eta$, potentially reaching the target performance with only a few high-impact interactions if the effective difficulty of the minimization problem is relatively low. 

Regarding efficiency, GBFS was almost consistently the most computationally expensive choice (highest SE). The relative efficiency difference between GFS and GR largely depended on where their respective forward or backward searches terminated, which is influenced by the effective difficulty of the minimization task for a given user and threshold $\eta$.

In summary, the choice between greedy removal and greedy selection methods depends heavily on the target performance level $\eta$. Greedy removal is better at maximizing data reduction under stricter performance constraints (high $\eta$). Greedy selection methods are better when the performance constraints are relaxed.

\subsubsection*{RQ2: What is the influence of user factors (esp. history size) on data reduction and computational cost?}
\label{sec:discussion_rq2}
Our second research question asked how user characteristics, particularly history size, influence data minimization. The stratified analysis presented in Tables 5 and 6 provides key insights into this question by examining performance across users grouped by history size bins. The datasets exhibit distinct distributions of user history lengths; Netflix generally features longer histories (e.g., 95th percentile \textasciitilde440 interactions) compared to ML25M (\textasciitilde270) and MSD (\textasciitilde160). Consequently, the bins used in our stratified analysis reflect these dataset-specific ranges.

Regarding the effect of history size, we observed distinct trends across the bins for the evaluated minimization algorithms. Users with longer histories generally exhibited a lower Minimization Ratio (MR), indicating that a larger fraction of their interaction data could be removed while meeting the performance constraint ($\eta$). This suggests that longer histories may contain more redundancy or information saturation, allowing for proportionally greater reduction compared to shorter, sparser histories (Tables 5 and 6). Correspondingly, the computational cost, measured by Sample Efficiency (SE), generally increases substantially for bins representing longer histories, particularly for the greedy algorithms (Tables 5 and 6). This is unsurprising, as longer histories present a larger search space. This highlights a characteristic: while users with the longest histories allow for the greatest proportional data reduction, minimizing their data is also the most computationally demanding.

\begin{figure}[ht]
    \centering
    \subfigure[EASE]{
        \includegraphics[width=0.48\columnwidth]{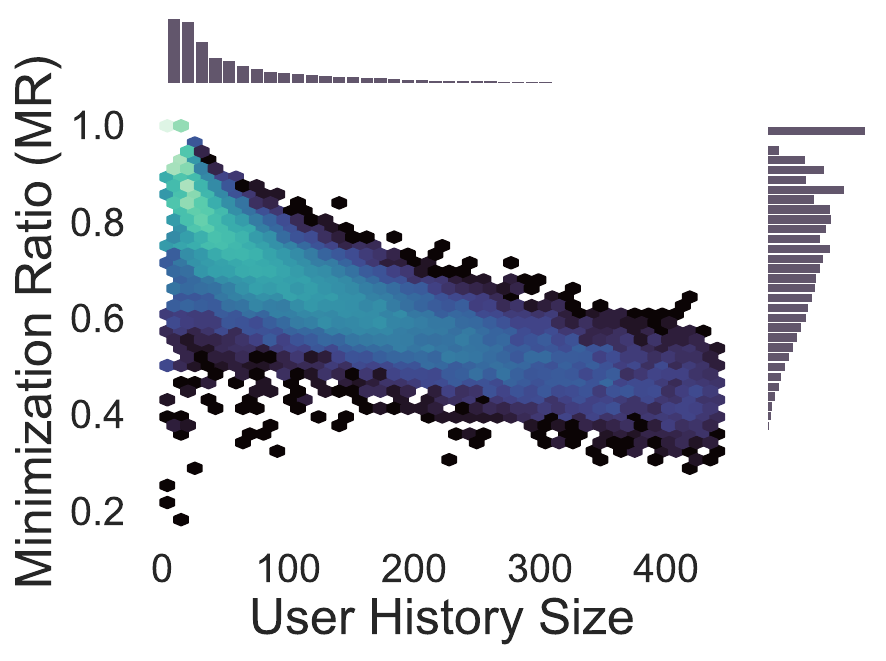}
        \label{fig:mr_ease}
    }%
    \hfill
    \subfigure[ItemKNN]{
        \includegraphics[width=0.48\columnwidth]{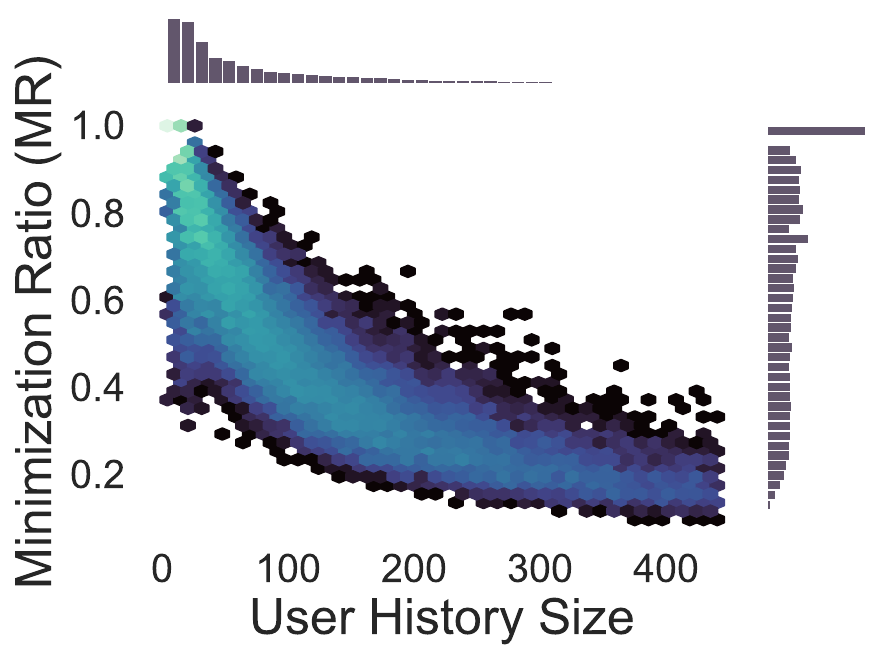}
        \label{fig:mr_itemknn}
    }
    \caption{Minimization Ratio (MR) vs. User History Size with GR on Netflix (for a target of $\eta=0.99$), comparing different recommendation models: EASE and ItemKNN. Note the significant variance in MR for users with similar history sizes. Lighter color indicates a higher density of users.}
    \label{fig:minimization-ratio-combined}
\end{figure}

Furthermore, our analysis indicates the existence of inherent differences in minimization potential even among users with similar history sizes. Examining the results within individual strata revealed notable variance: some users allowed for significantly greater reduction (lower MR) than others within the same history size bin while achieving the same relative performance $\eta$. This variance is visually confirmed by the vertical spread of data points for specific history sizes in Figure \ref{fig:minimization-ratio-combined}, which shows MR versus history size for the GR algorithm. We hypothesize that this variability represents differences in the inherent complexity or predictability of user preferences. Users exhibiting more diverse, niche, or rapidly evolving tastes might require a larger number of interactions to adequately inform the recommendation model, making their histories inherently harder to minimize without performance degradation compared to users with more predictable or mainstream preferences, even if their history sizes are similar.

Overall, our results confirm that user history size is a significant, observable factor influencing both the potential for and the cost of data minimization. Longer histories permit greater proportional reduction but incur higher computational costs. Additionally, the variance observed within strata suggests that latent factors likely related to preference complexity also contribute to minimization difficulty.
\section{Conclusion}\label{sec:conclusion}
This paper presented a comprehensive feasibility study focused on operationalizing the legal principle of data minimization for implicit feedback inference data within recommender systems. Our investigation characterized the performance-size relationship achievable in various minimization settings (\textbf{RQ1}) and explored the influence of user factors on reduction potential and cost (\textbf{RQ2}).

Addressing the relationship between recommendation performance and size (\textbf{RQ1}). Substantial data reduction is often achievable while maintaining high performance (e.g., $\eta=0.99$ and $\eta=0.98$). This directly addresses our title question, \textit{"What data is really necessary?"}, our results suggest that potentially large portions of user data currently collected, stored and used may not meet the strict legal definition of `necessary' for the purpose of presenting quality recommendations. However, the strictest requirement of perfect retention ($\eta=1.0$) proved largely impractical for a model like EASE, permitting very little data removal. This shows that the feasibility of minimization varies dramatically depending on the technical setting: the performance threshold ($\eta$), the recommendation model ($\recmod$), the evaluation metric ($\mathcal{R}$), and the ground truth estimation method all significantly influence the `effective difficulty' of the minimization task. This deep dependence on the technical setup highlights the inherent challenge in translating legal mandates like data minimization into practice; the necessity of data is effectively defined by specific, arguably arbitrary, technical choices. Set the `effective difficulty' too high, and most data appears essential; set it too low, and the majority of the data is deemed irrelevant. Achieving `necessity' therefore becomes a delicate balancing act, largely governed by this technical setting. Critically, these relatively arbitrary technical choices also shape our evaluation and development of minimization strategies. For instance, our work shows the \textbf{Greedy Removal} variant achieved superior data reduction at lower computational cost in high `effective difficulty' scenarios, whereas Greedy Selection approaches performed better in settings of lower difficulty. This demonstrates that the optimal minimization strategy itself is contingent on the chosen technical setting.
\begin{table*}[h]
\caption{Comparison of minimization algorithms based on Minimization Ratio (MR, \%) and Sample Efficiency (SE, \#) given a fixed performance threshold ($\eta=0.99$), stratified by user history size for the EASE model. Best MR highlighted in bold.}
\label{tab:strat-results-ease-99}
\centering
\sisetup{group-separator={,}, group-minimum-digits=4}
\begin{tabular}{@{}l l S[table-format=2.1] S[table-format=5.0] S[table-format=2.1] S[table-format=3.0] S[table-format=2.1] S[table-format=3.0] S[table-format=2.1] S[table-format=3.0] S[table-format=2.1] S[table-format=5.0] S[table-format=2.1] S[table-format=6.0] S[table-format=2.1, table-number-alignment=right, table-text-alignment=right] S[table-format=5.0]@{}}
\toprule
& & \multicolumn{2}{c}{\textbf{RS}} & \multicolumn{2}{c}{\textbf{LP}} & \multicolumn{2}{c}{\textbf{MP}} & \multicolumn{2}{c}{\textbf{EmbSim}} & \multicolumn{2}{c}{\textbf{GFS}} & \multicolumn{2}{c}{\textbf{GBFS(L=5)}} & \multicolumn{2}{c@{}}{\textbf{GR}} \\
\cmidrule(lr){3-4} \cmidrule(lr){5-6} \cmidrule(lr){7-8} \cmidrule(lr){9-10} \cmidrule(lr){11-12} \cmidrule(lr){13-14} \cmidrule(lr){15-16}
\textbf{Dataset} & \textbf{Hist. Bin} & {MR} & {SE} & {MR} & {SE} & {MR} & {SE} & {MR} & {SE} & {MR} & {SE} & {MR} & {SE} & {MR} & {SE} \\
\midrule
\multirow{5}{*}{Netflix}
 & (0, 88)    & 96.2 & 32    & 98.4 & 33    & 89.7 & 30    & 92.4 & 31    & 86.6 & 776    & 82.6 & 3538    & \textbf{77.8} & 368 \\
 & [88, 176)  & 95.0 & 121   & 98.5 & 126   & 88.0 & 112   & 88.5 & 113   & 87.0 & 8212   & 83.0 & 39464   & \textbf{64.3} & 5012 \\
 & [176, 264) & 94.8 & 205   & 98.8 & 214   & 87.5 & 190   & 86.7 & 188   & 88.5 & 23261  & 85.4 & 113425  & \textbf{56.3} & 16223 \\
 & [264, 352) & 95.0 & 289   & 99.0 & 301   & 87.4 & 266   & 86.0 & 262   & 89.7 & 45781  & 87.2 & 224632  & \textbf{50.4} & 34679 \\
 & [352, 440) & 95.1 & 373   & 99.1 & 389   & 86.9 & 341   & 85.4 & 336   & 90.6 & 76221  & 87.7 & 373495  & \textbf{46.0} & 60783 \\
\midrule
\multirow{5}{*}{ML25M}
 & (0, 54)    & 96.9 & 24    & 98.4 & 25    & 91.7 & 23    & 94.1 & 24    & 89.6 & 377    & 86.2 & 1655    & \textbf{84.7} & 138 \\
 & [54, 108)  & 95.3 & 74    & 98.1 & 77    & 87.8 & 69    & 98.4 & 70    & 87.2 & 3054   & 82.1 & 14379   & \textbf{74.0} & 1471 \\
 & [108, 162) & 94.8 & 126   & 98.4 & 131   & 86.2 & 115   & 86.9 & 116   & 86.0 & 8654   & 80.4 & 41252   & \textbf{66.6} & 4996 \\
 & [162, 216) & 94.9 & 178   & 98.8 & 185   & 85.5 & 160   & 85.4 & 160   & 85.3 & 17042  & 79.5 & 81563   & \textbf{61.5} & 10939 \\
 & [216, 270) & 94.5 & 228   & 98.7 & 238   & 85.3 & 192   & 83.5 & 201   & 84.6 & 28180  & 79.4 & 135783  & \textbf{57.1} & 19602 \\
\midrule
\multirow{5}{*}{MSD}
 & (0, 32)    & 98.0 & 23    & 98.4 & 23    & 98.7 & 23    & 98.5 & 23    & 88.0 & 272    & 86.9 & 1152    & \textbf{86.6} & 90 \\
 & [32, 64)   & 96.7 & 46    & 97.8 & 45    & 97.5 & 45    & 97.3 & 45    & 82.0 & 1035   & 79.3 & 4737    & \textbf{77.5} & 468 \\
 & [64, 96)   & 96.1 & 75    & 97.3 & 76    & 97.0 & 76    & 95.9 & 75    & 80.1 & 2916   & 76.5 & 13700   & \textbf{69.6} & 1625 \\
 & [96, 128)  & 95.8 & 106   & 97.0 & 107   & 96.7 & 107   & 94.5 & 105   & 81.8 & 5860   & 78.2 & 27907   & \textbf{64.1} & 3650 \\
 & [128, 160) & 95.6 & 137   & 97.1 & 139   & 96.7 & 139   & 93.1 & 134   & 84.0 & 9929   & 80.5 & 47741   & \textbf{60.5} & 6548 \\
\bottomrule
\end{tabular}
\end{table*}

\begin{table*}[h]
\caption{Comparison of minimization algorithms based on Minimization Ratio (MR, \%) and Sample Efficiency (SE, \#) given a fixed performance threshold ($\eta=0.99$), stratified by user history size for the ItemKNN model. Best MR highlighted in bold.}
\label{tab:strat-results-itemknn-99} 
\centering
\begin{tabular}{@{}l l r r r r r r r r r r r r r r@{}}
\toprule
& & \multicolumn{2}{c}{\textbf{RS}} & \multicolumn{2}{c}{\textbf{LP}} & \multicolumn{2}{c}{\textbf{MP}} & \multicolumn{2}{c}{\textbf{EmbSim}} & \multicolumn{2}{c}{\textbf{GFS}} & \multicolumn{2}{c}{\textbf{GBFS(L=5)}} & \multicolumn{2}{c@{}}{\textbf{GR}} \\
\cmidrule(lr){3-4} \cmidrule(lr){5-6} \cmidrule(lr){7-8} \cmidrule(lr){9-10} \cmidrule(lr){11-12} \cmidrule(lr){13-14} \cmidrule(lr){15-16}
\textbf{Dataset} & \textbf{Hist. Bin} & {MR} & {SE} & {MR} & {SE} & {MR} & {SE} & {MR} & {SE} & {MR} & {SE} & {MR} & {SE} & {MR} & {SE} \\
\midrule
\multirow{5}{*}{Netflix}
 & (0, 88)    & 91.3 & 31    & 95.2 & 32    & 89.4 & 30    & 88.2 & 30    & 73.3 & 692   & \textbf{68.0} & 3026   & \textbf{68.0} & 490 \\
 & [88, 176)  & 84.2 & 108   & 95.3 & 122   & 81.4 & 104   & 77.7 & 99    & 45.9 & 5634  & \textbf{39.1} & 24178  & 40.6 & 7032 \\
 & [176, 264) & 80.0 & 174   & 95.5 & 207   & 78.4 & 170   & 73.6 & 160   & 34.3 & 12971 & \textbf{28.9} & 55097  & 29.3 & 21591 \\
 & [264, 352) & 76.8 & 234   & 95.4 & 290   & 76.7 & 234   & 71.3 & 217   & 27.2 & 21178 & 22.8 & 89453  & \textbf{22.7} & 43980 \\
 & [352, 440) & 74.8 & 294   & 95.5 & 375   & 75.1 & 295   & 69.4 & 273   & 23.8 & 31370 & 19.7 & 131537 & \textbf{18.9} & 74331 \\
\midrule
\multirow{5}{*}{ML25M}
 & (0, 54)    & 75.4 & 19    & 92.3 & 23    & 84.9 & 21    & 78.9 & 20    & 50.5 & 243   & \textbf{44.2} & 934    & 52.9 & 291 \\
 & [54, 108)  & 59.6 & 47    & 92.6 & 72    & 78.9 & 62    & 71.0 & 56    & 25.0 & 1233  & \textbf{20.1} & 4677   & 30.3 & 2813 \\
 & [108, 162) & 49.7 & 67    & 92.4 & 123   & 75.2 & 100   & 66.5 & 89    & 15.9 & 2395  & \textbf{12.3} & 8788   & 20.8 & 8420 \\
 & [162, 216) & 43.8 & 83    & 93.0 & 174   & 74.0 & 139   & 63.0 & 118   & 11.4 & 3513  & \textbf{8.6} & 12637  & 15.6 & 17007 \\
 & [216, 270) & 38.9 & 94    & 92.9 & 224   & 73.0 & 176   & 61.3 & 148   & 9.5 & 4874  & \textbf{7.3} & 17596  & 13.1 & 28356 \\
\midrule
\multirow{5}{*}{MSD}
 & (0, 32)    & 94.0 & 22    & 96.0 & 23    & 94.7 & 22    & 92.8 & 22    & 78.6 & 259   & 75.9 & 1090   & \textbf{75.3} & 136 \\
 & [32, 64)   & 91.3 & 42    & 94.8 & 44    & 93.6 & 43    & 85.8 & 40    & 72.4 & 960   & 68.4 & 4313   & \textbf{61.0} & 696 \\
 & [64, 96)   & 88.9 & 70    & 94.3 & 74    & 93.0 & 73    & 79.2 & 62    & 68.6 & 2599  & 64.6 & 11951  & \textbf{50.2} & 2284 \\
 & [96, 128)  & 87.8 & 97    & 94.1 & 104   & 93.0 & 103   & 74.7 & 83    & 66.4 & 5028  & 62.6 & 23423  & \textbf{44.3} & 4873 \\
 & [128, 160) & 87.1 & 125   & 94.2 & 135   & 93.0 & 134   & 71.8 & 103   & 63.7 & 8197  & 59.6 & 38279  & \textbf{40.0} & 8534 \\
\bottomrule
\end{tabular}
\end{table*}

Analyzing user factors (\textbf{RQ2}), stratified results confirmed that history size significantly impacts reduction potential (lower MR for larger histories) and computational cost (higher SE for larger histories). Crucially, performance variance for users with similar history sizes indicates that latent factors beyond size, likely related to preference complexity, also influence minimization outcomes.

In conclusion, this work establishes the technical feasibility of substantial, per-user inference data reduction for recommender systems, while highlighting the complexities of operationalizing the data minimization principle. We find data necessity is not absolute but dependent on the technical setting (algorithms, metrics, goals) and user characteristics, challenging standardization. This motivates further research into practical minimization techniques and robust operationalizations of this fundamental legal principle.

\paragraph{Open Problems and Future Work}
In agreement with prior work~\cite{finck21,biega20} and informed by the findings of this study, several key areas warrant continued research. A central challenge remains the practical operationalization of data ‘necessity’. This is particularly true because recommender system performance is not assessed solely on ranking accuracy, but on a host of metrics across diverse dimensions~\cite{valcarce20,ricci15}, such as diversity~\cite{castells15} and serendipity~\cite{Kotkov16, Smets22}. This multi-objective evaluation landscape significantly complicates attempts to define what data is truly `necessary' for achieving an `adequate' overall system performance. Furthermore, the significant computational cost of current methods, underscored by our results, highlights an urgent need for more efficient and scalable minimization algorithms. Continued systematic assessment is also crucial to understand how feasibility and outcomes are influenced by contextual factors such as varying user characteristics, dataset properties, and model architectures. 

\section*{Acknowledgements}
This work was supported by Research Foundation — Flanders
(FWO) [11P2X24N to J. Leysen]. This work made use of \href{https://www.slices-ri.eu}{Slices-RI} infrastructure. We are grateful to the anonymous reviewers for their constructive suggestions, and to Lien Michiels for providing feedback on the draft of this paper.

\appendix

\section*{GenAI Usage Disclosure}
Pursuant to ACM policy, the authors disclose the use of Generative AI tools. Google Gemini's assistance was limited to improving the language, grammatical structure, and overall flow during the writing and editing phases of this manuscript. Additionally, GitHub Copilot was utilized as a coding assistant during software development and implementation. Generative AI was not employed for the conceptualization of the core research ideas, data analysis, or generation of the final results presented.

\bibliographystyle{ACM-Reference-Format}
\balance
\bibliography{ref}


\begin{thebibliography}{46}


\ifx \showCODEN    \undefined \def \showCODEN     #1{\unskip}     \fi
\ifx \showISBNx    \undefined \def \showISBNx     #1{\unskip}     \fi
\ifx \showISBNxiii \undefined \def \showISBNxiii  #1{\unskip}     \fi
\ifx \showISSN     \undefined \def \showISSN      #1{\unskip}     \fi
\ifx \showLCCN     \undefined \def \showLCCN      #1{\unskip}     \fi
\ifx \shownote     \undefined \def \shownote      #1{#1}          \fi
\ifx \showarticletitle \undefined \def \showarticletitle #1{#1}   \fi
\ifx \showURL      \undefined \def \showURL       {\relax}        \fi
\providecommand\bibfield[2]{#2}
\providecommand\bibinfo[2]{#2}
\providecommand\natexlab[1]{#1}
\providecommand\showeprint[2][]{arXiv:#2}

\bibitem[Bertin-Mahieux et~al\mbox{.}(2011)]%
        {bertin11}
\bibfield{author}{\bibinfo{person}{Thierry Bertin-Mahieux}, \bibinfo{person}{Daniel P~W Ellis}, \bibinfo{person}{Brian Whitman}, {and} \bibinfo{person}{Paul Lamere}.} \bibinfo{year}{2011}\natexlab{}.
\newblock \showarticletitle{{THE} {MILLION} {SONG} {DATASET}}. In \bibinfo{booktitle}{\emph{Proceedings of the 12th {International} {Society} for {Music} {Information} {Retrieval} {Conference} ({ISMIR} 2011))}}. \bibinfo{publisher}{University of Miami}, \bibinfo{address}{Miami, Florida, USA}, \bibinfo{pages}{591--596}.
\newblock
\urldef\tempurl%
\url{http://ismir2011.ismir.net/papers/OS6-1.pdf}
\showURL{%
\tempurl}


\bibitem[Biega et~al\mbox{.}(2020)]%
        {biega20}
\bibfield{author}{\bibinfo{person}{Asia~J. Biega}, \bibinfo{person}{Peter Potash}, \bibinfo{person}{Hal Daumé}, \bibinfo{person}{Fernando Diaz}, {and} \bibinfo{person}{Michèle Finck}.} \bibinfo{year}{2020}\natexlab{}.
\newblock \showarticletitle{Operationalizing the {Legal} {Principle} of {Data} {Minimization} for {Personalization}}. In \bibinfo{booktitle}{\emph{Proceedings of the 43rd {International} {ACM} {SIGIR} {Conference} on {Research} and {Development} in {Information} {Retrieval}}}. \bibinfo{publisher}{ACM}, \bibinfo{address}{Virtual Event China}, \bibinfo{pages}{399--408}.
\newblock
\showISBNx{978-1-4503-8016-4}
\href{https://doi.org/10.1145/3397271.3401034}{doi:\nolinkurl{10.1145/3397271.3401034}}


\bibitem[{Brasil}(2018)]%
        {LGPD2018}
\bibfield{author}{\bibinfo{person}{{Brasil}}.} \bibinfo{year}{2018}\natexlab{}.
\newblock \bibinfo{title}{{Lei Geral de Prote{\c{c}}{\~a}o de Dados Pessoais (LGPD)}}.
\newblock
\newblock
\shownote{Law No. 13.709 of August 14, 2018. Available at \url{http://www.planalto.gov.br/ccivil_03/_ato2015-2018/2018/lei/l13709.htm}}.


\bibitem[Castells et~al\mbox{.}(2015)]%
        {castells15}
\bibfield{author}{\bibinfo{person}{Pablo Castells}, \bibinfo{person}{Neil~J. Hurley}, {and} \bibinfo{person}{Saul Vargas}.} \bibinfo{year}{2015}\natexlab{}.
\newblock \showarticletitle{Novelty and Diversity in Recommender Systems}.
\newblock In \bibinfo{booktitle}{\emph{Recommender Systems Handbook}}, \bibfield{editor}{\bibinfo{person}{Francesco Ricci}, \bibinfo{person}{Lior Rokach}, {and} \bibinfo{person}{Bracha Shapira}} (Eds.). \bibinfo{publisher}{Springer US}, \bibinfo{address}{Boston, MA}, \bibinfo{pages}{881--918}.
\newblock
\showISBNx{978-1-4899-7637-6}
\href{https://doi.org/10.1007/978-1-4899-7637-6_26}{doi:\nolinkurl{10.1007/978-1-4899-7637-6_26}}


\bibitem[Cavoukian(2010)]%
        {cavoukian10}
\bibfield{author}{\bibinfo{person}{Ann Cavoukian}.} \bibinfo{year}{2010}\natexlab{}.
\newblock \showarticletitle{Privacy by design: the definitive workshop. A foreword}.
\newblock \bibinfo{journal}{\emph{Identity in the Information Society}} \bibinfo{volume}{3}, \bibinfo{number}{2} (\bibinfo{year}{2010}), \bibinfo{pages}{247--251}.
\newblock
\href{https://doi.org/10.1007/s12394-010-0062-y}{doi:\nolinkurl{10.1007/s12394-010-0062-y}}


\bibitem[Danezis et~al\mbox{.}(2014)]%
        {danezis14}
\bibfield{author}{\bibinfo{person}{George Danezis}, \bibinfo{person}{Josep Domingo-Ferrer}, \bibinfo{person}{Marit Hansen}, \bibinfo{person}{Jaap-Henk Hoepman}, \bibinfo{person}{Daniel Métayer}, \bibinfo{person}{Rodica Tirtea}, {and} \bibinfo{person}{Stefan Schiffner}.} \bibinfo{year}{2014}\natexlab{}.
\newblock \bibinfo{booktitle}{\emph{Privacy and Data Protection by Design - from Policy to Engineering}}.
\newblock \bibinfo{type}{{T}echnical {R}eport}. \bibinfo{institution}{European Union Agency for Network and Information Security (ENISA)}, \bibinfo{address}{Athens, Greece}.
\newblock
\href{https://doi.org/10.2824/38623}{doi:\nolinkurl{10.2824/38623}}


\bibitem[Eichinger and K\"{u}pper(2023)]%
        {eichinger23}
\bibfield{author}{\bibinfo{person}{Tobias Eichinger} {and} \bibinfo{person}{Axel K\"{u}pper}.} \bibinfo{year}{2023}\natexlab{}.
\newblock \showarticletitle{Distributed Data Minimization for Decentralized Collaborative Filtering Systems}. In \bibinfo{booktitle}{\emph{Proceedings of the 24th International Conference on Distributed Computing and Networking}} (Kharagpur, India) \emph{(\bibinfo{series}{ICDCN '23})}. \bibinfo{publisher}{Association for Computing Machinery}, \bibinfo{address}{New York, NY, USA}, \bibinfo{pages}{140–149}.
\newblock
\showISBNx{9781450397964}
\href{https://doi.org/10.1145/3571306.3571400}{doi:\nolinkurl{10.1145/3571306.3571400}}


\bibitem[Eichinger and Küpper(2024)]%
        {eichinger24}
\bibfield{author}{\bibinfo{person}{Tobias Eichinger} {and} \bibinfo{person}{Axel Küpper}.} \bibinfo{year}{2024}\natexlab{}.
\newblock \showarticletitle{On data minimization and anonymity in pervasive mobile-to-mobile recommender systems}.
\newblock \bibinfo{journal}{\emph{Pervasive and Mobile Computing}}  \bibinfo{volume}{103} (\bibinfo{year}{2024}), \bibinfo{pages}{101951}.
\newblock
\showISSN{1574-1192}
\href{https://doi.org/10.1016/j.pmcj.2024.101951}{doi:\nolinkurl{10.1016/j.pmcj.2024.101951}}


\bibitem[{European Parliament} and {Council of the European Union}(2016)]%
        {GDPR2016}
\bibfield{author}{\bibinfo{person}{{European Parliament}} {and} \bibinfo{person}{{Council of the European Union}}.} \bibinfo{year}{2016}\natexlab{}.
\newblock \bibinfo{title}{Regulation ({EU}) 2016/679 of the European Parliament and of the Council of 27 April 2016 on the protection of natural persons with regard to the processing of personal data and on the free movement of such data, and repealing Directive 95/46/EC (General Data Protection Regulation)}.
\newblock \bibinfo{howpublished}{Official Journal of the European Union L 119/1}.
\newblock
\urldef\tempurl%
\url{https://eur-lex.europa.eu/eli/reg/2016/679/oj}
\showURL{%
\tempurl}


\bibitem[Ferrari~Dacrema et~al\mbox{.}(2021)]%
        {dacrema21}
\bibfield{author}{\bibinfo{person}{Maurizio Ferrari~Dacrema}, \bibinfo{person}{Simone Boglio}, \bibinfo{person}{Paolo Cremonesi}, {and} \bibinfo{person}{Dietmar Jannach}.} \bibinfo{year}{2021}\natexlab{}.
\newblock \showarticletitle{A Troubling Analysis of Reproducibility and Progress in Recommender Systems Research}.
\newblock \bibinfo{journal}{\emph{ACM Trans. Inf. Syst.}} \bibinfo{volume}{39}, \bibinfo{number}{2}, Article \bibinfo{articleno}{20} (\bibinfo{date}{jan} \bibinfo{year}{2021}), \bibinfo{numpages}{49}~pages.
\newblock
\showISSN{1046-8188}
\href{https://doi.org/10.1145/3434185}{doi:\nolinkurl{10.1145/3434185}}


\bibitem[Ferrari~Dacrema et~al\mbox{.}(2019)]%
        {dacrema19}
\bibfield{author}{\bibinfo{person}{Maurizio Ferrari~Dacrema}, \bibinfo{person}{Paolo Cremonesi}, {and} \bibinfo{person}{Dietmar Jannach}.} \bibinfo{year}{2019}\natexlab{}.
\newblock \showarticletitle{Are we really making much progress? A worrying analysis of recent neural recommendation approaches}. In \bibinfo{booktitle}{\emph{Proceedings of the 13th ACM Conference on Recommender Systems}} (Copenhagen, Denmark) \emph{(\bibinfo{series}{RecSys '19})}. \bibinfo{publisher}{Association for Computing Machinery}, \bibinfo{address}{New York, NY, USA}, \bibinfo{pages}{101–109}.
\newblock
\showISBNx{9781450362436}
\href{https://doi.org/10.1145/3298689.3347058}{doi:\nolinkurl{10.1145/3298689.3347058}}


\bibitem[Finck and Biega(2021)]%
        {finck21}
\bibfield{author}{\bibinfo{person}{Michele Finck} {and} \bibinfo{person}{Asia~J. Biega}.} \bibinfo{year}{2021}\natexlab{}.
\newblock \showarticletitle{Reviving {Purpose} {Limitation} and {Data} {Minimisation} in {Data}-{Driven} {Systems}}.
\newblock \bibinfo{journal}{\emph{Technology and Regulation}}  \bibinfo{volume}{2021} (\bibinfo{year}{2021}), \bibinfo{pages}{44--61}.
\newblock
\href{https://doi.org/10.26116/TECHREG.2021.004}{doi:\nolinkurl{10.26116/TECHREG.2021.004}}


\bibitem[Harper and Konstan(2015)]%
        {harper16}
\bibfield{author}{\bibinfo{person}{F.~Maxwell Harper} {and} \bibinfo{person}{Joseph~A. Konstan}.} \bibinfo{year}{2015}\natexlab{}.
\newblock \showarticletitle{The MovieLens Datasets: History and Context}.
\newblock \bibinfo{journal}{\emph{ACM Trans. Interact. Intell. Syst.}} \bibinfo{volume}{5}, \bibinfo{number}{4}, Article \bibinfo{articleno}{19} (\bibinfo{date}{dec} \bibinfo{year}{2015}), \bibinfo{numpages}{19}~pages.
\newblock
\showISSN{2160-6455}
\href{https://doi.org/10.1145/2827872}{doi:\nolinkurl{10.1145/2827872}}


\bibitem[Hu et~al\mbox{.}(2008)]%
        {hu08}
\bibfield{author}{\bibinfo{person}{Yifan Hu}, \bibinfo{person}{Yehuda Koren}, {and} \bibinfo{person}{Chris Volinsky}.} \bibinfo{year}{2008}\natexlab{}.
\newblock \showarticletitle{Collaborative {Filtering} for {Implicit} {Feedback} {Datasets}}. In \bibinfo{booktitle}{\emph{2008 {Eighth} {IEEE} {International} {Conference} on {Data} {Mining}}}. \bibinfo{publisher}{IEEE}, \bibinfo{address}{Pisa, Italy}, \bibinfo{pages}{263--272}.
\newblock
\showISBNx{978-0-7695-3502-9}
\href{https://doi.org/10.1109/ICDM.2008.22}{doi:\nolinkurl{10.1109/ICDM.2008.22}}


\bibitem[Jannach et~al\mbox{.}(2018)]%
        {Jannach2018}
\bibfield{author}{\bibinfo{person}{Dietmar Jannach}, \bibinfo{person}{Lukas Lerche}, {and} \bibinfo{person}{Markus Zanker}.} \bibinfo{year}{2018}\natexlab{}.
\newblock \showarticletitle{Recommending Based on Implicit Feedback}.
\newblock In \bibinfo{booktitle}{\emph{Social Information Access: Systems and Technologies}}, \bibfield{editor}{\bibinfo{person}{Peter Brusilovsky} {and} \bibinfo{person}{Daqing He}} (Eds.). \bibinfo{publisher}{Springer International Publishing}, \bibinfo{address}{Cham}, \bibinfo{pages}{510--569}.
\newblock
\href{https://doi.org/10.1007/978-3-319-90092-6_14}{doi:\nolinkurl{10.1007/978-3-319-90092-6_14}}


\bibitem[J\"{a}rvelin and Kek\"{a}l\"{a}inen(2002)]%
        {jarvelin02}
\bibfield{author}{\bibinfo{person}{Kalervo J\"{a}rvelin} {and} \bibinfo{person}{Jaana Kek\"{a}l\"{a}inen}.} \bibinfo{year}{2002}\natexlab{}.
\newblock \showarticletitle{Cumulated gain-based evaluation of IR techniques}.
\newblock \bibinfo{journal}{\emph{ACM Trans. Inf. Syst.}} \bibinfo{volume}{20}, \bibinfo{number}{4} (\bibinfo{date}{oct} \bibinfo{year}{2002}), \bibinfo{pages}{422–446}.
\newblock
\showISSN{1046-8188}
\href{https://doi.org/10.1145/582415.582418}{doi:\nolinkurl{10.1145/582415.582418}}


\bibitem[Joachims et~al\mbox{.}(2007)]%
        {joachims07}
\bibfield{author}{\bibinfo{person}{Thorsten Joachims}, \bibinfo{person}{Laura Granka}, \bibinfo{person}{Bing Pan}, \bibinfo{person}{Helene Hembrooke}, \bibinfo{person}{Filip Radlinski}, {and} \bibinfo{person}{Geri Gay}.} \bibinfo{year}{2007}\natexlab{}.
\newblock \showarticletitle{Evaluating the accuracy of implicit feedback from clicks and query reformulations in Web search}.
\newblock \bibinfo{journal}{\emph{ACM Trans. Inf. Syst.}} \bibinfo{volume}{25}, \bibinfo{number}{2} (\bibinfo{date}{apr} \bibinfo{year}{2007}), \bibinfo{pages}{7–es}.
\newblock
\showISSN{1046-8188}
\href{https://doi.org/10.1145/1229179.1229181}{doi:\nolinkurl{10.1145/1229179.1229181}}


\bibitem[Kim and Suh(2019)]%
        {kim19}
\bibfield{author}{\bibinfo{person}{Daeryong Kim} {and} \bibinfo{person}{Bongwon Suh}.} \bibinfo{year}{2019}\natexlab{}.
\newblock \showarticletitle{Enhancing VAEs for collaborative filtering: flexible priors \& gating mechanisms}. In \bibinfo{booktitle}{\emph{Proceedings of the 13th ACM Conference on Recommender Systems}} (Copenhagen, Denmark) \emph{(\bibinfo{series}{RecSys '19})}. \bibinfo{publisher}{Association for Computing Machinery}, \bibinfo{address}{New York, NY, USA}, \bibinfo{pages}{403–407}.
\newblock
\showISBNx{9781450362436}
\href{https://doi.org/10.1145/3298689.3347015}{doi:\nolinkurl{10.1145/3298689.3347015}}


\bibitem[Kotkov et~al\mbox{.}(2016)]%
        {Kotkov16}
\bibfield{author}{\bibinfo{person}{Denis Kotkov}, \bibinfo{person}{Shuaiqiang Wang}, {and} \bibinfo{person}{Jari Veijalainen}.} \bibinfo{year}{2016}\natexlab{}.
\newblock \showarticletitle{A survey of serendipity in recommender systems}.
\newblock \bibinfo{journal}{\emph{Knowledge-Based Systems}}  \bibinfo{volume}{111} (\bibinfo{year}{2016}), \bibinfo{pages}{180--192}.
\newblock
\showISSN{0950-7051}
\href{https://doi.org/10.1016/j.knosys.2016.08.014}{doi:\nolinkurl{10.1016/j.knosys.2016.08.014}}


\bibitem[Larson et~al\mbox{.}(2017)]%
        {larson2017}
\bibfield{author}{\bibinfo{person}{Martha Larson}, \bibinfo{person}{Alessandro Zito}, \bibinfo{person}{Babak Loni}, {and} \bibinfo{person}{Paolo Cremonesi}.} \bibinfo{year}{2017}\natexlab{}.
\newblock \showarticletitle{Towards Minimal Necessary Data: The Case for Analyzing Training Data Requirements of Recommender Algorithms}. In \bibinfo{booktitle}{\emph{Proceedings of the Workshop on Responsible Recommendation at ACM RecSys'17 (FATREC'17)}}. \bibinfo{publisher}{Boise State University}, \bibinfo{address}{Boise, ID, USA}, \bibinfo{pages}{1--6}.
\newblock
\href{https://doi.org/10.18122/B2VX12}{doi:\nolinkurl{10.18122/B2VX12}}


\bibitem[Liang et~al\mbox{.}(2018)]%
        {liang18}
\bibfield{author}{\bibinfo{person}{Dawen Liang}, \bibinfo{person}{Rahul~G. Krishnan}, \bibinfo{person}{Matthew~D. Hoffman}, {and} \bibinfo{person}{Tony Jebara}.} \bibinfo{year}{2018}\natexlab{}.
\newblock \showarticletitle{Variational Autoencoders for Collaborative Filtering}. In \bibinfo{booktitle}{\emph{Proceedings of the 2018 World Wide Web Conference}} (Lyon, France) \emph{(\bibinfo{series}{WWW '18})}. \bibinfo{publisher}{International World Wide Web Conferences Steering Committee}, \bibinfo{address}{Republic and Canton of Geneva, CHE}, \bibinfo{pages}{689–698}.
\newblock
\showISBNx{9781450356398}
\href{https://doi.org/10.1145/3178876.3186150}{doi:\nolinkurl{10.1145/3178876.3186150}}


\bibitem[Lobel et~al\mbox{.}(2020)]%
        {lobel20}
\bibfield{author}{\bibinfo{person}{Sam Lobel}, \bibinfo{person}{Chunyuan Li}, \bibinfo{person}{Jianfeng Gao}, {and} \bibinfo{person}{Lawrence Carin}.} \bibinfo{year}{2020}\natexlab{}.
\newblock \showarticletitle{RaCT: Toward Amortized Ranking-Critical Training for Collaborative Filtering}. In \bibinfo{booktitle}{\emph{Proceedings of the 8th International Conference on Learning Representations (ICLR'20)}}. International Conference on Learning Representations (ICLR), \bibinfo{address}{Addis Ababa, Ethiopia}.
\newblock


\bibitem[Michiels et~al\mbox{.}(2022)]%
        {michiels22}
\bibfield{author}{\bibinfo{person}{Lien Michiels}, \bibinfo{person}{Robin Verachtert}, {and} \bibinfo{person}{Bart Goethals}.} \bibinfo{year}{2022}\natexlab{}.
\newblock \showarticletitle{{RecPack}: {An}(other) {Experimentation} {Toolkit} for {Top}-{N} {Recommendation} using {Implicit} {Feedback} {Data}}. In \bibinfo{booktitle}{\emph{Proceedings of the 16th {ACM} {Conference} on {Recommender} {Systems}}}. \bibinfo{publisher}{ACM}, \bibinfo{address}{Seattle WA USA}, \bibinfo{pages}{648--651}.
\newblock
\showISBNx{978-1-4503-9278-5}
\href{https://doi.org/10.1145/3523227.3551472}{doi:\nolinkurl{10.1145/3523227.3551472}}


\bibitem[{National Assembly of the Republic of Korea}(2011)]%
        {PIPA11}
\bibfield{author}{\bibinfo{person}{{National Assembly of the Republic of Korea}}.} \bibinfo{year}{2011}\natexlab{}.
\newblock \bibinfo{title}{Personal Information Protection Act (PIPA)}.
\newblock \bibinfo{howpublished}{Act No. 10465, Mar. 29, 2011, as amended. English versions available through the Personal Information Protection Commission (PIPC) Korea or Korea Law Translation Center.}
\newblock
\urldef\tempurl%
\url{https://www.law.go.kr/LSW/eng/engLsSc.do?menuId=2&section=lawNm&query=Personal+Information+Protection+Act&x=0&y=0#liBgPtn}
\showURL{%
\tempurl}


\bibitem[{National People's Congress of the People's Republic of China}(2021)]%
        {PIPL_Stanford_2021}
\bibfield{author}{\bibinfo{person}{{National People's Congress of the People's Republic of China}}.} \bibinfo{year}{2021}\natexlab{}.
\newblock \bibinfo{title}{Personal Information Protection Law of the People's Republic of China}.
\newblock \bibinfo{howpublished}{English translation by DigiChina Project, Stanford Cyber Policy Center}.
\newblock
\urldef\tempurl%
\url{https://digichina.stanford.edu/work/translation-personal-information-protection-law-of-the-peoples-republic-of-china-effective-nov-1-2021/}
\showURL{%
\tempurl}
\newblock
\shownote{Adopted by the Standing Committee of the National People's Congress on August 20, 2021. Effective November 1, 2021.}.


\bibitem[Niu et~al\mbox{.}(2023)]%
        {niu23}
\bibfield{author}{\bibinfo{person}{Xi Niu}, \bibinfo{person}{Ruhani Rahman}, \bibinfo{person}{Xiangcheng Wu}, \bibinfo{person}{Zhe Fu}, \bibinfo{person}{Depeng Xu}, {and} \bibinfo{person}{Riyi Qiu}.} \bibinfo{year}{2023}\natexlab{}.
\newblock \showarticletitle{Leveraging Uncertainty Quantification for Reducing Data for Recommender Systems}. In \bibinfo{booktitle}{\emph{2023 IEEE International Conference on Big Data (BigData)}}. \bibinfo{publisher}{IEEE}, \bibinfo{address}{New York, NY, USA}, \bibinfo{pages}{352--359}.
\newblock
\href{https://doi.org/10.1109/BigData59044.2023.10386790}{doi:\nolinkurl{10.1109/BigData59044.2023.10386790}}


\bibitem[{Parliament of Canada}(2000)]%
        {PIPEDA00}
\bibfield{author}{\bibinfo{person}{{Parliament of Canada}}.} \bibinfo{year}{2000}\natexlab{}.
\newblock \bibinfo{title}{Personal Information Protection and Electronic Documents Act (S.C. 2000, c. 5)}.
\newblock
\urldef\tempurl%
\url{https://laws-lois.justice.gc.ca/eng/acts/p-8.6/}
\showURL{%
\tempurl}


\bibitem[{Parliament of the Republic of South Africa}(2013)]%
        {POPIA13}
\bibfield{author}{\bibinfo{person}{{Parliament of the Republic of South Africa}}.} \bibinfo{year}{2013}\natexlab{}.
\newblock \bibinfo{title}{Protection of Personal Information Act 4 of 2013 (POPIA)}.
\newblock \bibinfo{howpublished}{Assented to 19 November 2013. Most operative provisions commenced 1 July 2020.}
\newblock
\urldef\tempurl%
\url{https://www.gov.za/documents/protection-personal-information-act-popia}
\showURL{%
\tempurl}


\bibitem[Rendle et~al\mbox{.}(2009)]%
        {Rendle09}
\bibfield{author}{\bibinfo{person}{Steffen Rendle}, \bibinfo{person}{Christoph Freudenthaler}, \bibinfo{person}{Zeno Gantner}, {and} \bibinfo{person}{Lars Schmidt-Thieme}.} \bibinfo{year}{2009}\natexlab{}.
\newblock \showarticletitle{BPR: Bayesian personalized ranking from implicit feedback}. In \bibinfo{booktitle}{\emph{Proceedings of the Twenty-Fifth Conference on Uncertainty in Artificial Intelligence}} (Montreal, Quebec, Canada) \emph{(\bibinfo{series}{UAI '09})}. \bibinfo{publisher}{AUAI Press}, \bibinfo{address}{Arlington, Virginia, USA}, \bibinfo{pages}{452–461}.
\newblock
\showISBNx{9780974903958}


\bibitem[Rendle et~al\mbox{.}(2022)]%
        {rendle22}
\bibfield{author}{\bibinfo{person}{Steffen Rendle}, \bibinfo{person}{Walid Krichene}, \bibinfo{person}{Li Zhang}, {and} \bibinfo{person}{Yehuda Koren}.} \bibinfo{year}{2022}\natexlab{}.
\newblock \showarticletitle{Revisiting the Performance of iALS on Item Recommendation Benchmarks}. In \bibinfo{booktitle}{\emph{Proceedings of the 16th ACM Conference on Recommender Systems}} (Seattle, WA, USA) \emph{(\bibinfo{series}{RecSys '22})}. \bibinfo{publisher}{Association for Computing Machinery}, \bibinfo{address}{New York, NY, USA}, \bibinfo{pages}{427–435}.
\newblock
\showISBNx{9781450392785}
\href{https://doi.org/10.1145/3523227.3548486}{doi:\nolinkurl{10.1145/3523227.3548486}}


\bibitem[Rendle and Zhang(2023)]%
        {rendle23}
\bibfield{author}{\bibinfo{person}{Steffen Rendle} {and} \bibinfo{person}{Li Zhang}.} \bibinfo{year}{2023}\natexlab{}.
\newblock \showarticletitle{On Reducing User Interaction Data for Personalization}.
\newblock \bibinfo{journal}{\emph{ACM Trans. Recomm. Syst.}} \bibinfo{volume}{1}, \bibinfo{number}{3}, Article \bibinfo{articleno}{14} (\bibinfo{date}{aug} \bibinfo{year}{2023}), \bibinfo{numpages}{28}~pages.
\newblock
\href{https://doi.org/10.1145/3600097}{doi:\nolinkurl{10.1145/3600097}}


\bibitem[Ricci et~al\mbox{.}(2010)]%
        {ricci15}
\bibfield{author}{\bibinfo{person}{Francesco Ricci}, \bibinfo{person}{Lior Rokach}, \bibinfo{person}{Bracha Shapira}, {and} \bibinfo{person}{Paul~B. Kantor}.} \bibinfo{year}{2010}\natexlab{}.
\newblock \bibinfo{booktitle}{\emph{Recommender Systems Handbook} (\bibinfo{edition}{1st} ed.)}.
\newblock \bibinfo{publisher}{Springer-Verlag}, \bibinfo{address}{Berlin, Heidelberg}.
\newblock
\showISBNx{0387858199}


\bibitem[Sachdeva et~al\mbox{.}(2022)]%
        {sachdeva22OnSampling}
\bibfield{author}{\bibinfo{person}{Noveen Sachdeva}, \bibinfo{person}{Carole-Jean Wu}, {and} \bibinfo{person}{Julian McAuley}.} \bibinfo{year}{2022}\natexlab{}.
\newblock \showarticletitle{On Sampling Collaborative Filtering Datasets}. In \bibinfo{booktitle}{\emph{Proceedings of the Fifteenth ACM International Conference on Web Search and Data Mining}} (Virtual Event, AZ, USA) \emph{(\bibinfo{series}{WSDM '22})}. \bibinfo{publisher}{Association for Computing Machinery}, \bibinfo{address}{New York, NY, USA}, \bibinfo{pages}{842–850}.
\newblock
\showISBNx{9781450391320}
\href{https://doi.org/10.1145/3488560.3498439}{doi:\nolinkurl{10.1145/3488560.3498439}}


\bibitem[Sarwar et~al\mbox{.}(2001)]%
        {sarwar01}
\bibfield{author}{\bibinfo{person}{Badrul~M. Sarwar}, \bibinfo{person}{George Karypis}, \bibinfo{person}{Joseph~A. Konstan}, {and} \bibinfo{person}{John~T. Riedl}.} \bibinfo{year}{2001}\natexlab{}.
\newblock \showarticletitle{Item-Based Collaborative Filtering Recommendation Algorithms}. In \bibinfo{booktitle}{\emph{Proceedings of the 10th International Conference on World Wide Web (WWW '01)}}. \bibinfo{publisher}{ACM}, \bibinfo{address}{New York, NY, USA}, \bibinfo{pages}{285--295}.
\newblock
\href{https://doi.org/10.1145/371920.372071}{doi:\nolinkurl{10.1145/371920.372071}}


\bibitem[Shanmugam et~al\mbox{.}(2022)]%
        {shanmugam22}
\bibfield{author}{\bibinfo{person}{Divya Shanmugam}, \bibinfo{person}{Fernando Diaz}, \bibinfo{person}{Samira Shabanian}, \bibinfo{person}{Michele Finck}, {and} \bibinfo{person}{Asia Biega}.} \bibinfo{year}{2022}\natexlab{}.
\newblock \showarticletitle{Learning to Limit Data Collection via Scaling Laws: A Computational Interpretation for the Legal Principle of Data Minimization}. In \bibinfo{booktitle}{\emph{Proceedings of the 2022 ACM Conference on Fairness, Accountability, and Transparency}} (Seoul, Republic of Korea) \emph{(\bibinfo{series}{FAccT '22})}. \bibinfo{publisher}{Association for Computing Machinery}, \bibinfo{address}{New York, NY, USA}, \bibinfo{pages}{839–849}.
\newblock
\showISBNx{9781450393522}
\href{https://doi.org/10.1145/3531146.3533148}{doi:\nolinkurl{10.1145/3531146.3533148}}


\bibitem[Shehzad and Jannach(2023)]%
        {shehzad23}
\bibfield{author}{\bibinfo{person}{Faisal Shehzad} {and} \bibinfo{person}{Dietmar Jannach}.} \bibinfo{year}{2023}\natexlab{}.
\newblock \showarticletitle{Everyone's a Winner! On Hyperparameter Tuning of Recommendation Models}. In \bibinfo{booktitle}{\emph{Proceedings of the 17th ACM Conference on Recommender Systems}} (Singapore, Singapore) \emph{(\bibinfo{series}{RecSys '23})}. \bibinfo{publisher}{Association for Computing Machinery}, \bibinfo{address}{New York, NY, USA}, \bibinfo{pages}{652–657}.
\newblock
\showISBNx{9798400702419}
\href{https://doi.org/10.1145/3604915.3609488}{doi:\nolinkurl{10.1145/3604915.3609488}}


\bibitem[Shenbin et~al\mbox{.}(2020)]%
        {shenbin20}
\bibfield{author}{\bibinfo{person}{Ilya Shenbin}, \bibinfo{person}{Anton Alekseev}, \bibinfo{person}{Elena Tutubalina}, \bibinfo{person}{Valentin Malykh}, {and} \bibinfo{person}{Sergey~I. Nikolenko}.} \bibinfo{year}{2020}\natexlab{}.
\newblock \showarticletitle{RecVAE: A New Variational Autoencoder for Top-N Recommendations with Implicit Feedback}. In \bibinfo{booktitle}{\emph{Proceedings of the 13th International Conference on Web Search and Data Mining}} (Houston, TX, USA) \emph{(\bibinfo{series}{WSDM '20})}. \bibinfo{publisher}{Association for Computing Machinery}, \bibinfo{address}{New York, NY, USA}, \bibinfo{pages}{528–536}.
\newblock
\showISBNx{9781450368223}
\href{https://doi.org/10.1145/3336191.3371831}{doi:\nolinkurl{10.1145/3336191.3371831}}


\bibitem[Smets et~al\mbox{.}(2022)]%
        {Smets22}
\bibfield{author}{\bibinfo{person}{Annelien Smets}, \bibinfo{person}{Lien Michiels}, \bibinfo{person}{Toine Bogers}, {and} \bibinfo{person}{Lennart Björneborn}.} \bibinfo{year}{2022}\natexlab{}.
\newblock \showarticletitle{Serendipity in Recommender Systems Beyond the Algorithm: a Feature Repository and Experimental Design}. In \bibinfo{booktitle}{\emph{Joint Workshop on Interfaces and Human Decision Making at RecSys 2022}} \emph{(\bibinfo{series}{CEUR Workshop Proceedings}, Vol.~\bibinfo{volume}{3222})}. \bibinfo{publisher}{CEUR Workshop Proceedings}, \bibinfo{pages}{46--66}.
\newblock
\urldef\tempurl%
\url{https://ceur-ws.org/Vol-3222/paper4.pdf}
\showURL{%
\tempurl}


\bibitem[{State of California}(2020)]%
        {CPRA2020}
\bibfield{author}{\bibinfo{person}{{State of California}}.} \bibinfo{year}{2020}\natexlab{}.
\newblock \bibinfo{title}{{California Privacy Rights Act of 2020 (CPRA)}}.
\newblock
\newblock
\shownote{Approved by voters November 3, 2020. Amends the California Consumer Privacy Act (CCPA). Codified in Cal. Civ. Code § 1798.100 et seq. Available at \url{https://leginfo.legislature.ca.gov/faces/codes_displaySection.xhtml?sectionNum=1798.140.&lawCode=CIV}}.


\bibitem[Steck(2019a)]%
        {steck19_b}
\bibfield{author}{\bibinfo{person}{Harald Steck}.} \bibinfo{year}{2019}\natexlab{a}.
\newblock \bibinfo{title}{Collaborative Filtering via High-Dimensional Regression}.
\newblock
\showeprint[arxiv]{1904.13033}
\urldef\tempurl%
\url{http://arxiv.org/abs/1904.13033}
\showURL{%
\tempurl}
\newblock
\shownote{arXiv:1904.13033 [cs]}.


\bibitem[Steck(2019b)]%
        {steck19_a}
\bibfield{author}{\bibinfo{person}{Harald Steck}.} \bibinfo{year}{2019}\natexlab{b}.
\newblock \showarticletitle{Embarrassingly Shallow Autoencoders for Sparse Data}. In \bibinfo{booktitle}{\emph{The World Wide Web Conference}} (San Francisco, CA, USA) \emph{(\bibinfo{series}{WWW '19})}. \bibinfo{publisher}{Association for Computing Machinery}, \bibinfo{address}{New York, NY, USA}, \bibinfo{pages}{3251–3257}.
\newblock
\showISBNx{9781450366748}
\href{https://doi.org/10.1145/3308558.3313710}{doi:\nolinkurl{10.1145/3308558.3313710}}


\bibitem[{The Federal Assembly of the Swiss Confederation}(2020)]%
        {SwissFADP20}
\bibfield{author}{\bibinfo{person}{{The Federal Assembly of the Swiss Confederation}}.} \bibinfo{year}{2020}\natexlab{}.
\newblock \bibinfo{title}{Federal Act on Data Protection (FADP) of 25 September 2020}.
\newblock \bibinfo{howpublished}{Status as of 1 September 2023. SR 235.1}.
\newblock
\urldef\tempurl%
\url{https://www.fedlex.admin.ch/eli/cc/2022/491/en}
\showURL{%
\tempurl}


\bibitem[{United Kingdom Parliament}(2018)]%
        {UKGDPR18}
\bibfield{author}{\bibinfo{person}{{United Kingdom Parliament}}.} \bibinfo{year}{2018}\natexlab{}.
\newblock \bibinfo{title}{The UK General Data Protection Regulation (UK GDPR)}.
\newblock \bibinfo{howpublished}{Based on Regulation (EU) 2016/679, as it forms part of domestic law by virtue of section 3 of the European Union (Withdrawal) Act 2018 and as amended.}
\newblock
\urldef\tempurl%
\url{https://www.legislation.gov.uk/eur/2016/679/contents}
\showURL{%
\tempurl}


\bibitem[Valcarce et~al\mbox{.}(2020)]%
        {valcarce20}
\bibfield{author}{\bibinfo{person}{Daniel Valcarce}, \bibinfo{person}{Alejandro Bellogín}, \bibinfo{person}{Javier Parapar}, {and} \bibinfo{person}{Pablo Castells}.} \bibinfo{year}{2020}\natexlab{}.
\newblock \showarticletitle{Assessing ranking metrics in top-{N} recommendation}.
\newblock \bibinfo{journal}{\emph{Information Retrieval Journal}} \bibinfo{volume}{23}, \bibinfo{number}{4} (\bibinfo{date}{Aug.} \bibinfo{year}{2020}), \bibinfo{pages}{411--448}.
\newblock
\showISSN{1386-4564, 1573-7659}
\href{https://doi.org/10.1007/s10791-020-09377-x}{doi:\nolinkurl{10.1007/s10791-020-09377-x}}


\bibitem[Verachtert et~al\mbox{.}(2022)]%
        {verachtert2022forgetting}
\bibfield{author}{\bibinfo{person}{Robin Verachtert}, \bibinfo{person}{Lien Michiels}, {and} \bibinfo{person}{Bart Goethals}.} \bibinfo{year}{2022}\natexlab{}.
\newblock \showarticletitle{Are We Forgetting Something? Correctly Evaluate a Recommender System With an Optimal Training Window}. In \bibinfo{booktitle}{\emph{Proceedings of the Perspectives on the Evaluation of Recommender Systems Workshop (PERSPECTIVES 2022)}} (Seattle, WA, USA, September 22nd, 2022). \bibinfo{publisher}{CEUR Workshop Proceedings}.
\newblock
\showISSN{1613-0073}
\urldef\tempurl%
\url{http://ceur-ws.org}
\showURL{%
\tempurl}


\bibitem[Vinagre and Jorge(2012)]%
        {vinagre2012forgetting}
\bibfield{author}{\bibinfo{person}{João Vinagre} {and} \bibinfo{person}{Alípio~Mário Jorge}.} \bibinfo{year}{2012}\natexlab{}.
\newblock \showarticletitle{Forgetting mechanisms for scalable collaborative filtering}.
\newblock \bibinfo{journal}{\emph{Journal of the Brazilian Computer Society}} \bibinfo{volume}{18}, \bibinfo{number}{4} (\bibinfo{year}{2012}), \bibinfo{pages}{271--282}.
\newblock
\showISSN{1678-4804}
\href{https://doi.org/10.1007/s13173-012-0077-3}{doi:\nolinkurl{10.1007/s13173-012-0077-3}}


\end{thebibliography}

\end{document}